\documentclass{article}

 \usepackage[preprint]{neurips_2026}

\usepackage[utf8]{inputenc} % allow utf-8 input
\usepackage[T1]{fontenc}    % use 8-bit T1 fonts
\usepackage{hyperref}       % hyperlinks
\usepackage{url}            % simple URL typesetting
\usepackage{booktabs}       % professional-quality tables
\usepackage{amsfonts}       % blackboard math symbols
\usepackage{nicefrac}       % compact symbols for 1/2, etc.
\usepackage{microtype}      % microtypography
\usepackage{xcolor}         % colors
\usepackage{listings}       % code listings
\usepackage{graphicx}
\usepackage{subcaption}
\usepackage{wrapfig}
\usepackage{amsmath}
\usepackage{amssymb}
\usepackage{mathtools}
\usepackage{amsthm}
\usepackage{multirow}
\usepackage{makecell}
\usepackage{enumitem}

\usepackage[capitalize,noabbrev]{cleveref}

\theoremstyle{plain}

\theoremstyle{definition}

\theoremstyle{remark}

\usepackage[textsize=tiny]{todonotes}

\title{Revisiting the Shape Convention of Transformer Language Models}

\author{%
  Feng-Ting Liao\textsuperscript{1}\thanks{Equal contribution.}\hspace{0.55em}\thanks{Correspondence to: \texttt{\{ft.liao, meng-hsi.chen\}@mtkresearch.com}} \And
  Guan-Ting Yi\textsuperscript{1, 2}\footnotemark[1] \And
  Tzu-Quan Lin\textsuperscript{1, 2}\footnotemark[1] \And
  Meng-Hsi Chen\textsuperscript{1}\footnotemark[2] \AND
  Da-shan Shiu\textsuperscript{1} \\
  \\
  \textsuperscript{1}MediaTek Research \quad \textsuperscript{2}National Taiwan University
}

\begin{document}

\maketitle

\begin{abstract}
  % Dense Transformer language models have largely adhered to one consistent architectural shape: each layer consists of an attention module followed by a feed-forward network (FFN) with a narrow--wide--narrow MLP, allocating most parameters to the MLP at expansion ratios between 2 and 4.
  % Motivated by recent results that residual wide--narrow--wide (hourglass) MLPs offer superior function approximation capabilities, we revisit the long-standing MLP shape convention in Transformer, challenging the necessity of the narrow--wide--narrow design.
  % To study this, we develop a Transformer variant that replaces the conventional FFN with a deeper hourglass-shaped FFN, comprising a stack of hourglass sub-MLPs connected by residual pathways.
  % We posit that a deeper but lighter hourglass FFN can serve as a competitive alternative to the conventional FFN, and that parameters saved by using a lighter hourglass FFN can be more effectively utilized, such as by enlarging model hidden dimensions under fixed budgets.
  % We confirm these through empirical validations across model scales: hourglass FFNs outperform conventional FFNs up to 400M and achieve comparable performance at larger scales to 1B parameters; hourglass FFN variants with reduced FFN and increased attention parameters show consistent improvements over conventional configurations at matched budgets.
  % Together, these findings shed new light on recent work and prompt a rethinking of the narrow-wide-narrow MLP convention and the balance between attention and FFN towards efficient and expressive modern language models.
  The architectural shape of dense Transformers has remained remarkably stable: narrow-wide-narrow feed-forward networks (FFNs) consume most non-embedding parameters.
  Motivated by theoretical and empirical evidences that residual wide-narrow-wide (hourglass) MLPs remain expressive despite bottlenecks, we revisit whether this architectural convention is necessary for dense language models.
  We study Hourglass Transformers, which replace the conventional FFN with residual stacks of hourglass sub-MLPs and use hourglass attention to decouple residual-stream width from attention width.
  This exposes a practical depth-width trade-off: compressing the FFN intermediate dimension allows wider hidden states and fewer layers at matched parameter budgets.
  % Across 113M to 8B parameters, Hourglass Transformers remain a comparable language-modeling and downstream quality frontier while being $8.7\%$ more compute efficient under matched average accuracy at 906M, 3B, and 8B. 
  Across model scales from 113M to 8B parameters, Hourglass Transformers achieve language-modeling and downstream performance comparable to conventional Transformers, while improving training compute efficiency by $8.7\%$ at matched average downstream accuracy across the 906M, 3B, and 8B scales.
  % The 8B Hourglass model, after long context extension, outperforms matched conventional baseline on long-context tasks across 4k--64k contexts.
  After long-context extension, the 8B Hourglass model also outperforms its matched conventional baseline across 4k–64k context lengths.
  At 64k context, the reduced attention layer count lowers both computation and KV-cache requirements, yielding up to $1.93\times$ faster token decoding and $50\%$ lower KV-cache memory at the 1B scale.
  These results identify hourglass structures as a practical architecture-efficiency alternative for compute- and latency-conscious Transformer design. 
\end{abstract}

% \begin{figure}[th!]
%     \centering
%     \includegraphics[width=0.9\linewidth]{figures/val_loss_vs_flops_small_v1.png}
%     \caption{\textbf{Performance frontiers of Transformers with hourglass (wide-narrow-wide) versus conventional (narrow-wide-narrow) FFNs} \citep{touvron2023llama}. We revisit the shape convention of Transformer by replacing the narrow-wide-narrow FFN with a hourglass FFN, composing stacks of wide-narrow-wide sub-MLPs connected by residuals. We observe that Hourglass FFNs achieve comparable performance to the conventional design up to 1B parameters. Here we also show a conventional variant trained based on OLMo-2 architecture. Only the non-embedding parameters are accounted for the FLOPs.}
%     \label{fig:scaling_plot}
%      \vspace{-4mm}
% \end{figure}

\vspace{-4mm}
\section{Introduction}
\label{sec:introduction}
\vspace{-2mm}

Despite rapid advances in scale and training methodology, the architectural shape of dense Transformer language models has remained remarkably stable since early scaling studies \citep{kaplan2020scalinglawsneurallanguage}. Modern models consistently adopt a narrow–wide–narrow multilayer perceptron (MLP) in feed-forward network (FFN), expanding the model dimension $d_{\text{model}}$ to an intermediate width $d_h$ before projecting back, with expansion ratios typically fixed between 2 and 4 \citep{vaswani2017attentionneed,kaplan2020scalinglawsneurallanguage}. This design choice has become a de facto standard across contemporary dense LLMs \citep{touvron2023llama,qwen2025qwen25technicalreport,gemmateam2025gemma3technicalreport,olmo20252olmo2furious}. At the same time, the FFN dominates parameter allocation in Transformers, accounting for a majority of model parameters relative to attention. As a result, the MLP shape implicitly determines how capacity is distributed between depth, width, and attention.

\begin{figure*}[ht!]
    \centering
    \includegraphics[width=0.7\textwidth]{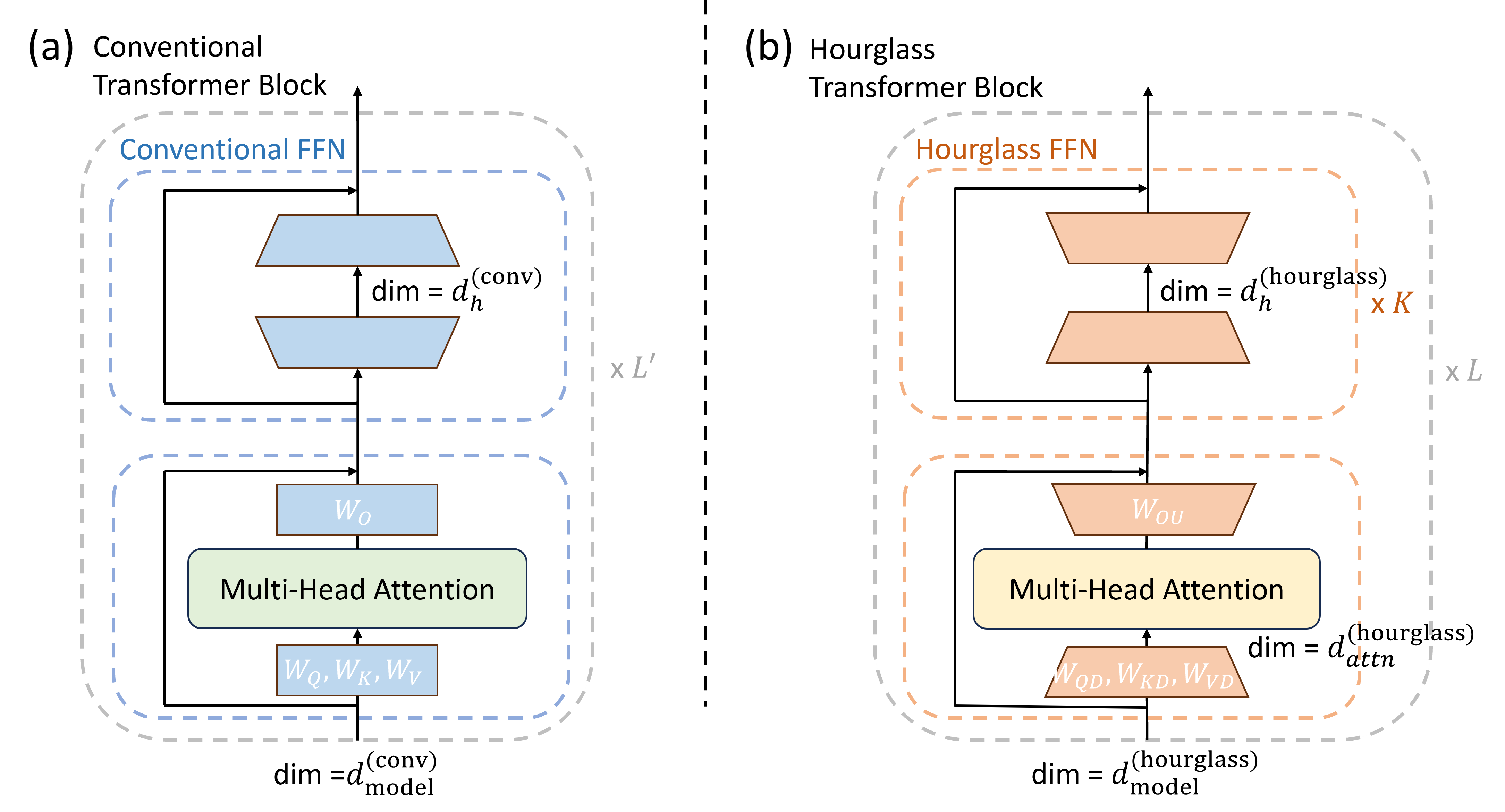}
    \caption{\textbf{Overview: revisiting the shape convention of Transformer through studying the relaxation of shape in FFN and attention.} Inspired by \citep{pmlr-v235-liu24am,chen2025rethinkingshapeconventionmlp}, we compare conventional and hourglass variants of Transformer architectures. (a) Conventional Transformer Block with $L'$ layers, consisting of an attention module and a conventional FFN with a narrow-wide-narrow MLP. (b) Hourglass Transformer Block with $L$ layers, consisting of an hourglass attention module and an hourglass FFN with $K$ hourglass-shaped MLP sub-blocks. We explore the design space by tuning parameters such as $d_{\text{model}}$, $d_h$, $K$, and $L$, allowing the Hourglass layer count $L$ to differ from the baseline $L'$ while fixing the Hourglass $d^{(\text{hourglass})}_{attn}=d^{(\text{conv})}_{\text{model}}$.}
    \label{fig:hourglass_block}
    \vspace{-5mm}
\end{figure*}

While the shape convention in Transformer is effective, we raise a fundamental question: \emph{is the narrow–wide–narrow MLP shape in FFN essential, or merely one convenient instantiation of a residual-compatible transformation?}
Recent results suggest that this convention may be unnecessarily restrictive.
Theoretical studies have shown that residual MLPs with wide–narrow–wide (hourglass) structures can act as optimal function approximators, including universality results for networks with extremely narrow intermediate layers \citep{NEURIPS2018_03bfc1d4,pmlr-v235-liu24am}.
From an expressivity perspective, such hourglass architectures motivate deeper residual compositions at fixed parameter budgets, a setting known to favor rapid growth in linear regions and representational complexity \citep{ICML-2019-HaninR,10.1007/978-3-032-09645-6_10}.
Moreover, operating in a wider residual embedding space before and after bottleneck transformations is consistent with the intuition behind Cover's theorem and random feature methods: high-dimensional representations can make useful distinctions easier to express \cite{cover1965covertheorm,Rahimi2007}.
These complementary arguments are not a substitute for language-model evidence, but they motivate treating bottleneck residual transformations as viable candidates rather than degenerate approximations.

Empirical evidence has also begun to challenge conventional MLP shape assumptions.
Motivated by high-dimensional representation learning, \citet{chen2025rethinkingshapeconventionmlp} demonstrates that hourglass-shaped MLPs outperform standard designs in generative vision models.
Related work further shows that expert-routed FFNs with hourglass MLPs in Mixture-of-Experts Transformers can surpass dense baselines \citep{wang-etal-2024-scaling} at scale. More broadly, the principle of allocating parameters through wide–narrow–wide pathways has proven effective in convolutional architectures \cite{BMVC2016_87,10.1007/978-3-030-58580-8_40} and parameter-efficient methods such as LoRA \cite{hu2022lora}.
Together, these results indicate that hourglass-shaped transformations constitute a strong and general class of building blocks, motivating a re-examination of shape conventions in Transformer.

In this work, we revisit the shape convention of Transformer by substituting both the conventional FFN and attention module with hourglass variants. The hourglass FFN replaces the single narrow–wide–narrow MLP with a stack of $K$ wide–narrow–wide sub-MLPs connected by residuals; the hourglass attention module projects the embedding to a lower dimension before applying attention, allowing sequence mixing to operate below the widened residual dimension.
Stacking $K$ residual sub-blocks deepens the compositional structure of the FFN, a regime where representational complexity can grow rapidly with depth \citep{10.1007/978-3-032-09645-6_10}. Practically, the inversion gives an additional architectural knob: FFN bottleneck width, residual-stream width, and the number of sequential Transformer layers can be traded against one another under a fixed parameter or FLOP budget.
Furthermore, recent depth–width studies have shown that, at fixed parameter count, deeper Transformers achieve lower perplexity than wider ones, but the benefit of additional layers saturates rapidly while the latency cost grows linearly in depth~\citep{petty-etal-2024-impact}. The hourglass FFN shifts this balance by making it possible to use a wider $d_{\text{model}}$ with fewer Transformer layers, the regime where latency savings are greatest and depth returns may already be diminished. This leads us to ask: \emph{can inverting the FFN shape exploit the depth–width trade-off more favorably, reducing training compute and latency while preserving quality?}

To answer this question, we construct Transformer variants with hourglass FFN and hourglass attention modules (collectively referred to as Hourglass Transformers) and conduct experiments across model scales from 113M to 8B parameters.\footnote{\href{https://github.com/mtkresearch/Hourglass_LM}{github.com/mtkresearch/Hourglass\_LM}}
% \footnote{Code will be released upon publication.}
Controlled studies at 113M isolate the contribution of FFN parameterization, while design-space sweeps at 403M/906M examine how hourglass FFN/Attn together support fewer-layer, wider-residual configurations. 
Scaling to 906M, 3B, and 8B parameters, we evaluate the complete Hourglass architecture recipe against standard Transformer baselines: Hourglass Transformers remain competitive across language modeling and question answering benchmarks, while improving training compute efficiency by $8.7\%$ at matched average downstream accuracy. 
At a 64k-token context, the reduced attention-layer count lowers KV-cache memory by $50.0\%$, $35.3\%$, and $31.3\%$ and improves token decoding speed by $1.93\times$, $1.52\times$, and $1.43\times$ for the 906M, 3B, and 8B models, respectively.
To test whether fewer attention layers compromise context extension, we additionally apply YaRN to the 8B model. 
Notably, Hourglass obtains lower long-context language modeling perplexity and retrieval and processing tasks than the conventional baseline at every evaluated length from 4k to 64k tokens.

To summarize, our contributions are as follows:
% \begin{itemize}
\begin{itemize}[noitemsep, topsep=0pt, leftmargin=*]
    \item We systematically investigate whether the narrow--wide--narrow FFN convention in Transformers can be replaced by an inverted, hourglass shape. We introduce Hourglass Transformers, which pair wide--narrow--wide FFN sub-blocks with hourglass attention modules for efficient sequence mixing in widened residual streams.
    \item Through controlled experiments at 113M and design-space exploration at 403M/906M, we show that hourglass FFNs are competitive with conventional FFNs while enabling a depth--width trade-off: fewer Transformer layers and wider residual representations at matched parameter budgets.
    \item We demonstrate that the complete Hourglass recipe persists through scales, yielding a competitive quality frontier with more efficient training FLOPs, faster token decoding at 64k context; an improved long-context performance at 8B.
\end{itemize}

\section{Background and Related Work}
\label{sec:background}
\vspace{-2mm}
\subsection{The Narrow-Wide-Narrow FFN Convention}
\vspace{-2mm}

In Transformer architectures, the FFN serves as a position-wise non-linear transformation applied independently to each token representation. Formally, given a hidden state $z\in \mathbb{R}^{d_\text{model}}$, the conventional MLP with residual in FFN computes
\begin{equation}
    \text{FFN}(z)=z+W_2 \sigma(W_1 \text{norm}(z))
\end{equation}
where $\sigma(\cdot)$ is a non-linear activation function, $\text{norm}(\cdot)$ is a normalization function, $W_1\in \mathbb{R}^{d_h\times d_\text{model}}$, $W_2\in \mathbb{R}^{d_\text{model}\times d_h}$, and $d_h > d_\text{model}$. This expansion–projection structure has been retained across most modern dense LLMs \cite{touvron2023llama,qwen2025qwen25technicalreport,gemmateam2025gemma3technicalreport}.

Beyond functional form, the FFN plays a central role in determining the overall shape of a Transformer. It accounts for a majority of parameters and floating-point operations per layer, often exceeding those of the attention module. As a result, the choice of the shape expansion ratio $d_h/d_\text{model}$, together with the model dimension $d_{\text{model}}$ and the number of block layers $L$, implicitly governs how capacity is allocated across width, depth, and attention. While prior work has explored variations in activation functions \citep{shazeer2020gluvariantsimprovetransformer}, normalization strategies \citep{xiong2020on,jiang2023prermsnorm}, parameter sharing \citep{Lan2020ALBERT,li-etal-2022-ode}, and depth-width trade-off \citep{petty-etal-2024-impact}, the overall narrow–wide–narrow shape of the FFN and its interaction with Transformer scaling \citep{kaplan2020scalinglawsneurallanguage} have remained largely unchanged.

% \subsection{Related work}

\subsection{Hourglass and Bottleneck Residual Architectures}
\vspace{-2mm}

Theoretical analyses of hourglass\footnote{Prior work often uses \textit{bottleneck} to describe the wide-narrow-wide shape; we use \textit{hourglass} interchangeably.} MLPs with residual connections have established universal approximation guarantees. With sufficient depth, \citep{NEURIPS2018_03bfc1d4} shows that residual networks with extremely narrow hidden layers retain universal approximation capability, while \citep{pmlr-v235-liu24am} further shows that residual hourglass MLPs with narrow constant width can achieve optimal approximation of Lebesgue-integrable functions.

Empirically, several efficient architectures have explored hourglass or bottleneck structures with residual connections. In computer vision, U-Net \citep{10.1007/978-3-319-24574-4_28}, MobileNet \citep{10.1007/978-3-030-58580-8_40}, and Wide ResNets \citep{BMVC2016_87} use bottleneck structures or separable convolutions to improve the parameter-compute ratio. These examples illustrate that carefully structured bottlenecks can reduce computational cost while preserving, or even enhancing, representational power through increased depth or improved gradient flow. In language modeling, low-rank and bottleneck-inspired techniques have primarily appeared in parameter-efficient methods. LoRA \citep{hu2022lora,10.5555/3692070.3693369} introduces low-rank projections within linear layers to reduce trainable parameters during fine-tuning, effectively imposing a bottleneck on weight updates. Mixture-of-Experts (MoE) \citep{shazeer2017outrageously,mu2025comprehensivesurveymixtureofexpertsalgorithms} scales capacity without a proportional increase in inference cost by replacing dense FFNs with sparse expert networks.

Earlier scaling-law experiments reported that hourglass MLP in dense Transformer FFNs underperformed their narrow-wide-narrow counterparts \citep{kaplan2020scalinglawsneurallanguage,petty-etal-2024-impact}.
More recent work on generative vision models paints a different picture: residual-connected hourglass MLPs that iteratively refine representations in expanded feature spaces can outperform conventional MLPs \citep{chen2025rethinkingshapeconventionmlp}.
Whereas prior work primarily studies general MLP properties or sparse expert settings, our work studies hourglass MLPs as dense Transformer FFNs, enabling a reconsideration of model shape and parameter and compute allocation between attention and FFN across language-model scales.

% More recently, complementary work by \citep{chen2025rethinkingshapeconventionmlp} shows that residual-connected hourglass MLPs, which iteratively refine representations in expanded feature spaces, outperform standard MLPs on generative vision tasks. 
% Furthermore, complementary analyses based on linear region counting \citep{pmlr-v80-zhang18i,ICML-2019-HaninR,10.1007/978-3-032-09645-6_10} further demonstrate that residual connections enable later layers to introduce decision boundaries that depend jointly on multiple earlier representations. Work by \citep{10.1007/978-3-032-09645-6_10} shows that skip connections, unlike plain feedforward networks, where hyperplanes are composed sequentially, allow linear regions to accumulate in a combinatorial manner across layers. 
% This effect persists even when intermediate layers are narrow, suggesting that bottleneck residual networks can achieve rapid growth in functional complexity despite limited per-layer width.

% Notably, \citep{chen2025rethinkingshapeconventionmlp} complementarily shows that iterative refinement of representation at expanded dimension with residual connected hourglass MLP is more effective than conventional MLPs on generative vision tasks.

\vspace{-2mm}

\section{Hourglass Transformer Architecture}
\label{sec:method}
\vspace{-2mm}

We define Hourglass Transformers by replacing the conventional wide-expansion FFN with a stack of residual bottleneck MLP sub-blocks (Hourglass FFN) and by decoupling residual-stream width from attention width through an hourglass attention module (Hourglass Attention) (See \autoref{fig:hourglass_block}).

\subsection{Network Architecture}
Our Hourglass Transformer follows the conventional LLaMA style Transformer backbone~\citep{grattafiori2024llama3herdmodels} consisting of input embeddings, stacked model layers $L$, and a final output projection.

\paragraph{Hourglass Transformer Layer}
Each layer consists of an attention module followed by an FFN. The attention module performs global information aggregation across the sequence, while the FFN performs local feature refinement within each token. Both components are wrapped with residual connections and layer normalization.

Specifically, given the input to the $l$-th layer $\mathbf{z}^{(\ell)}$, the intermediate representation $\mathbf{u}^{(\ell)}$ is computed via the attention mechanism (Attn): $\mathbf{u}^{(\ell)} = \mathbf{z}^{(\ell)} + \text{Attn}(\text{norm}(\mathbf{z}^{(\ell)}))$.
Then, we set $\mathbf{h}_0^{(\ell)} = \mathbf{u}^{(\ell)}$ as the input to the FFN.

\paragraph{Hourglass Feed-Forward Network}
The Hourglass FFN refines the representation through $K$ stacked hourglass-shaped MLP sub-blocks. For $i = 0, \dots, K-1$:
\begin{align}
    \mathbf{h}_{i+1}^{(\ell)} & = \mathbf{h}_i^{(\ell)} + \text{MLP}_i(\mathbf{h}_i^{(\ell)})
\end{align}
where $\text{MLP}_i$ denotes the $i$-th hourglass sub-block. Finally, the output of the layer is $\mathbf{z}^{(\ell+1)} = \mathbf{h}_K^{(\ell)}$.

While the convention expands the hidden dimension $d_{\text{model}}$ to a wider $d_{h}$, the Hourglass FFN utilizes a compression-expansion structure with a bottleneck dimension $d_h < d_{\text{model}}$. Each sub-block $\text{MLP}_i$ consists of a down-projection $W_d^{(i)} \in \mathbb{R}^{d_h \times d_{\text{model}}}$, a non-linear activation $\sigma$, and an up-projection $W_u^{(i)} \in \mathbb{R}^{d_{\text{model}} \times d_h}$. Formally:
\begin{equation}
    \text{MLP}_i(\mathbf{x}) = W_u^{(i)} \sigma(W_d^{(i)} \text{norm}( \mathbf{x}))
\end{equation}
This structure allows independent control over the FFN's depth ($K$) and width ($d_h$).

In practice, we implemented the hourglass MLP following \citep{grattafiori2024llama3herdmodels} using the SwiGLU activation function.
Specifically, each sub-block $\text{MLP}_i$ consists of two down-projection matrices $W_{d1}^{(i)}, W_{d2}^{(i)} \in \mathbb{R}^{d_h \times d_{\text{model}}}$ and one up-projection matrix $W_u^{(i)} \in \mathbb{R}^{d_{\text{model}} \times d_h}$. That is
\begin{equation}
    \text{MLP}_i(\mathbf{x}) = W_u^{(i)} (\text{SiLU}(W_{d1}^{(i)} \bar{\mathbf{x}}) \odot (W_{d2}^{(i)} \bar{\mathbf{x}}))
\end{equation}
where $\bar{\mathbf{x}} = \text{norm}(\mathbf{x})$ and $\odot$ denotes element-wise multiplication.

\paragraph{Hourglass Attention}
Transformer with hourglass FFN enables a rethinking of the attention module width. Narrowing each FFN sub-block makes it possible to explore wider residual representations under a fixed parameter budget. However, increasing $d_{\text{model}}$ also increases the computational cost of attention if attention is performed directly in the residual dimension.

We propose Hourglass Attention to project the input to a lower dimensional space, perform attention, and project back to the original dimension, while keeping the computational cost of attention the same as the conventional attention. Specifically, the input $\mathbf{z}$ is first projected to a lower dimensional space by down-projection matrices $W_{QD}, W_{KD}, W_{VD} \in \mathbb{R}^{d_{\text{attn}} \times d_{\text{model}}}$, where $d_{\text{attn}} < d_{\text{model}}$. Then, the attention mechanism is performed on the projected key, query, and value features. Finally, the output of the attention mechanism is projected back to the original dimension by an up-projection output matrix $W_{OU} \in \mathbb{R}^{d_{\text{model}} \times d_{\text{attn}}}$. Formally:
\begin{equation}
    \text{Attn}(\mathbf{z}) = W_{OU} \text{softmax}\left(\frac{(W_{QD} \mathbf{z}) (W_{KD} \mathbf{z})^T}{\sqrt{d_{\text{attn}}}} \right) W_{VD} \mathbf{z},
\end{equation}
where QK-norm, RoPE, and multi-head settings are omitted for brevity. In practice, we implement the multi-head attention by splitting the projected key, query, and value features into multiple heads, and perform attention on each head independently.
Note that when $d_{\text{attn}} = d_{\text{model}}$, the Hourglass Attention reduces to the conventional multi-head attention.
In our main scaling experiments, we set $d_{\text{attn}}$ equal to the $d_{\text{model}}$ of the corresponding conventional baseline, so attention operates in the same-width attention space while the Hourglass residual stream can be wider.
Appendix~\ref{sec:attn_reduction} provides a sensitivity check where this attention dimension is further reduced.

% \subsection{Transformer Blocks with Hourglass Shape}
\subsection{The Hourglass Design Space}
\vspace{-2mm}

The hourglass shape exposes an explicit parameter trade-off between residual width, attention width, FFN bottleneck width, FFN depth, and Transformer depth.
Ignoring embeddings, normalization parameters, and biases, the dominant parameter count is $N \approx L\left(4d_{\text{model}}d_{\text{attn}} + 3K d_h d_{\text{model}}\right)$,
where the first term corresponds to the query, key, value, and output attention projections, and the second term corresponds to the three SwiGLU FFN projections in each of the $K$ hourglass sub-blocks.
For a token budget $T$ and sequence length $S$, the corresponding dominant forward-pass FLOP estimate is
\begin{align}
    C_{\text{fwd}} \approx T \left(L \left(8d_{\text{model}}d_{\text{attn}} + 2S d_{\text{attn}} + 6K d_h d_{\text{model}}\right) + 2d_{\text{model}}n_{\text{vocab}}\right),
\end{align}
where $n_{\text{vocab}}$ is the vocabulary size; Appendix~\ref{app:train_flops} gives the detailed forward--backward FLOP accounting used in our experiments.
To achieve strong performance at a given parameter and compute budget, we need to balance the design parameters $K$, $L$, $d_h$, $d_{\text{attn}}$, and $d_{\text{model}}$.
This design introduces two fundamental shifts in resource allocation:

\paragraph{Trading FFN Expansion and Layers for Residual Width.}
The Hourglass FFN ($d_h < d_{\text{model}}$ inside each sub-block) decouples FFN expressivity from a single large expansion ratio. Under a fixed parameter budget, this creates room to widen the residual stream and adjust the number of sequential Transformer layers without requiring attention to operate at the full residual dimension. Section~\ref{sec:experiments} evaluates whether this trade-off preserves quality while reducing training FLOPs and token decode latency.

\paragraph{Trading FFN Width for Depth.}
Conventional FFNs rely on extreme width within one MLP transformation. In contrast, the Hourglass FFN leverages depth by stacking $K$ residual sub-blocks within a single Transformer layer. This increases the sequence of non-linear transformations without requiring each sub-block to use a wide intermediate dimension. The resulting topology is a practical way to test whether deeper bottleneck transformations can substitute for the conventional per-layer FFN expansion in language models.

\paragraph{Inference Efficiency.} Considering the aforementioned parametrized trade-offs, the forward KV memory and compute costs of Hourglass Transformers for inference can be approximated as 
\begin{equation}
    M_{\text{KV}}(S;g)=\frac{2\beta L S d_{\text{attn}}}{g}
    \qquad
    \text{and}\quad
    t_{\text{decode}}(S;g)
    =t_{\text{fixed}}+\frac{4L S d_{\text{attn}}}{g\,\mathrm{BW}_{\text{eff}}},
\end{equation}
where $g$ is the per-layer KV-compression factor given a selected sequence mixing module such as MHA or GQA, $t_\text{fixed}$ is the fixed decoding time overhead, $\beta$ is the bytes per cached element, and $BW_{\text{eff}}$ is the effective GPU memory bandwidth. Section~\ref{app:inference_cost_analysis} provides detailed inference efficiency analysis, and Section~\ref{sec:latency} validates the theoretical inference efficiency gain with empirical measurements against the convetional baselines.

\section{Experiments}
\label{sec:experiments}
\vspace{-2mm}

% We evaluate the proposed hourglass FFN architecture against baseline Transformer-based LMs across multiple model scales.
% Rather than merely benchmarking performance, our experiments serve as a test for the conventional narrow-wide-narrow shape.
% We first use controlled settings to isolate the FFN-side effect, then evaluate the complete Hourglass architecture recipe as a depth--width trade-off against standard Transformer baselines.
We evaluate whether the conventional narrow--wide--narrow Transformer FFN can be replaced by an hourglass-shaped alternative without sacrificing language-model quality.
We first use controlled settings to isolate the FFN-side effect, then evaluate the complete Hourglass architecture as a depth--width trade-off against standard Transformer baselines.
% Our experiments first isolate the effect of the FFN parameterization under controlled settings, then study the broader Hourglass Transformer design space in which FFN bottleneck width, residual-stream width, attention width, and Transformer depth are traded under matched parameter budgets.
% Finally, we evaluate the resulting Hourglass architecture recipe against standard Transformer baselines across model scales.

\vspace{-1mm}
\subsection{Experimental Setup}
\label{sec:experimental_setup}
\vspace{-2mm}

\paragraph{Baselines.}
We compare against published dense Transformer configurations spanning 113M, 403M, 906M, 3B, and 8B parameters (see \autoref{tab:baseline_and_hourglass_configs}). These baselines follow LLaMA/OLMo-style architectures~\citep{touvron2023llama,olmo20252olmo2furious} and use standard shapes from prior scaling studies and open model reports~\citep{NEURIPS2020_1457c0d6_gpt3,touvron2023llama,grattafiori2024llama3herdmodels}. We treat these published shapes as established, reproducible reference points for conventional dense Transformers, not as baselines tuned around our method. As a robustness check, Appendix~\ref{sec:appendix_model_efficiency} reports conventional FFN-ratio variants at 403M and a reduced-depth parameter-matched variant at 906M; these controls shift the quality--compute trade-off but do not uniformly improve over the published-shape baselines.

%     \vspace{-2mm}
% \begin{itemize}
% % \begin{itemize}[noitemsep, topsep=0pt]
% \vspace{-2mm}
%     \item \textbf{Standard Transformer:} Canonical Transformer-based LMs following the LLaMA architecture~\cite{touvron2023llama,olmo20252olmo2furious}, serving as the primary conventional baselines for parameter efficiency comparisons ~\cite{gu2024mambalineartimesequencemodeling,sun2025learninglearntesttime}.
%     \vspace{-2mm}
% \item \textbf{Conventional (OLMo-2):} We take configuration of a state-of-the-art open-weights 1B-scale model~\cite{olmo20252olmo2furious}, used as a high-water mark for performance. OLMo2 differs from the standard Transformer baselines at the order of normalization, where the layer normalization taking place after the attention and the MLP but before the residuals.
% \vspace{-1mm}
% \end{itemize}

% The detailed configurations of all baselines are summarized in \autoref{tab:baseline_LLM_config}.

\vspace{-2mm}
\paragraph{Hourglass Transformers.}
For each baseline size, we construct a corresponding variant by replacing the conventional FFN with the Hourglass FFN and using Hourglass Attention. For fair comparison, $d_{\text{attn}}$ at each parameter size is chosen to match the $d_{\text{model}}$ of the corresponding conventional baseline.
Model configurations are chosen with approximately matched total parameter budgets, staying within $0.5\%$ of the baseline parameter count while keeping training FLOPs closely matched or lower.

\vspace{-2mm}
\paragraph{Training and Evaluation Details.}
All models are trained on the same dataset and tokenized using identical preprocessing pipelines.
Optimization uses AdamW with a cosine learning rate schedule. The details of experiment settings can be found in Appendix \ref{sec:experiment_setting_details}, where we follow general settings from \citep{NEURIPS2020_1457c0d6_gpt3,hoffmann2022trainingcomputeoptimallargelanguage,olmo20252olmo2furious}.
Following the convention in \citep{kaplan2020scalinglawsneurallanguage,hoffmann2022trainingcomputeoptimallargelanguage}, we report only non-embedding parameters in model-size comparisons. For training efficiency, we report training FLOPs per forward-backward pass, following \citep{hoffmann2022trainingcomputeoptimallargelanguage,mcleish2025gemstones}; the calculation script is provided in Appendix~\ref{app:train_flops}.
For inference, we report relative token-decode latency compared with the corresponding baseline model.
For evaluation, we report average validation loss and validation perplexity (PPL) over validation sets. Section~\ref{sec:main_scales} additionally reports downstream task performance at larger model scales.
\vspace{-2mm}

% \subsection{FFN Controls at 113M}
\subsection{A Controlled Test of FFN Shape}
\label{sec:ffn_efficiency_ablations}
\vspace{-2mm}

\begin{wraptable}{r}{0.45\textwidth}
    \vspace{-4mm}
    \centering
    \caption{\textbf{Hourglass FFNs vs. Conventional FFNs under fixed $d_\text{model}=768$ and $L=12$ (113M).} All models use 113M total parameters, with 28M attention parameters and 85M FFN parameters.}
    \label{tab:fix_attn}
    \resizebox{\linewidth}{!}{
        \begin{tabular}{lcc|cc}
            \toprule
            Model        & $d_{\text{h}}$ & $K$ & Val Loss $\downarrow$ & Val PPL $\downarrow$ \\
            \midrule
            Conventional & 3072           & 1   & 3.766                 & 43.19                \\
            \midrule
            \multirow{4}{*}{Hourglass}
                         & 614            & 5   & 3.749                 & 42.46                \\
                         & 512            & 6   & 3.752                 & 42.61                \\
                         & 384            & 8   & 3.742                 & 42.16                \\
                         & 307            & 10  & 3.747                 & 42.38                \\
            \bottomrule
        \end{tabular}
    }
    \vspace{-2mm}
\end{wraptable}

To isolate the contribution of the Hourglass FFN from changes in attention capacity and layer count, we evaluate 113M models where $d_{\text{model}}=768$ and $L=12$ are kept identical, ensuring that the attention parameter count and train FLOPs are the same across all variants (see \autoref{tab:fix_attn}).
We vary the Hourglass FFN depth $K$ and adjust the intermediate dimension $d_h$ accordingly, while keeping the total model size fixed at 113M parameters.
Under this controlled setting, the Hourglass FFN Transformer consistently achieves lower validation loss and perplexity than the conventional baseline.
This provides controlled evidence that the hourglass FFN parameterization can be competitive even when attention capacity is held fixed.
As an additional sanity check, Appendix~\ref{sec:appendix_bigram} reports a no-attention Bigram LM comparison, where the Hourglass FFN also remains comparable to the conventional FFN.
\vspace{-2mm}

\subsection{Effect of Larger \texorpdfstring{$d_{\text{model}}$}{d-model} in Hourglass Transformers}
\label{sec:frozen_lm_head}
\vspace{-2mm}

The Hourglass architecture compresses the FFN intermediate dimension, allowing configurations that widen $d_{\text{model}}$ under a matched parameter budget.
A wider hidden dimension increases the representational capacity of every residual-stream component, i.e., attention projections, layer norms, and the LM head.
To study this effect in isolation, we conduct a frozen LM head experiment on 403M models, which controls for the additional training FLOPs that a larger LM head would otherwise introduce.

\begin{figure}[ht!]
    \centering
    \begin{minipage}[t]{0.42\textwidth}
        \vspace{0pt}
        \centering
        \includegraphics[width=\linewidth]{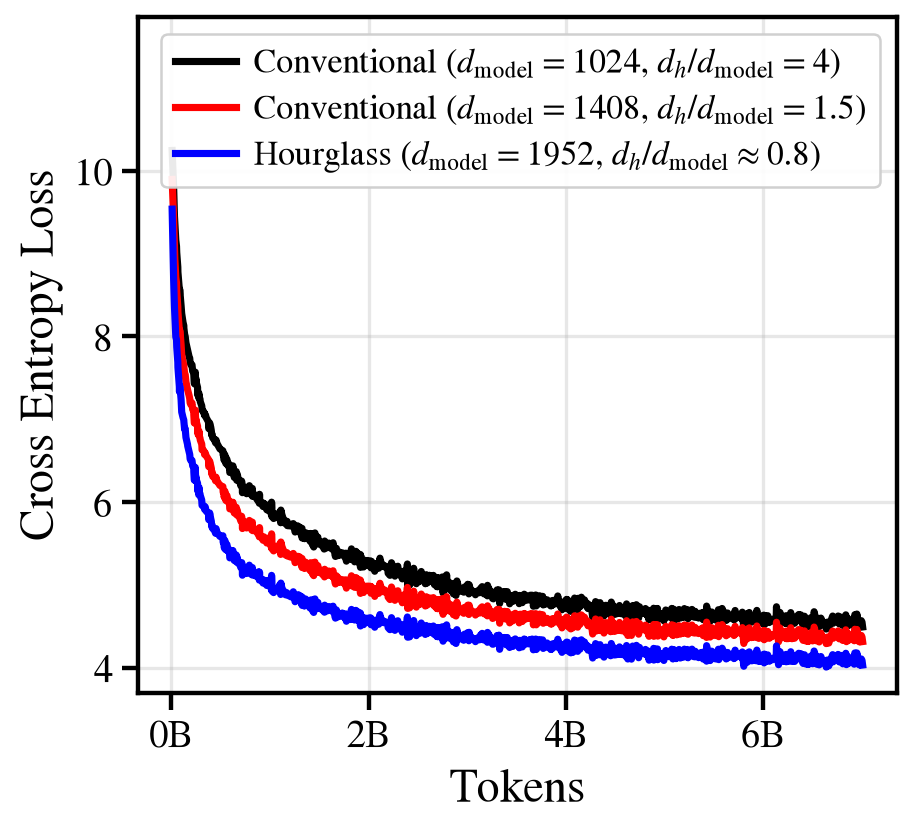}
        \captionsetup{hypcap=false}
        \captionof{figure}{\textbf{Frozen LM head Experiment at 403M.}
            The experiment probes the benefit of a larger $d_{\text{model}}$: the Hourglass model achieves the lowest loss, and the wider conventional variant outperforms the standard one, consistent with the gradient bottleneck hypothesis \citep{godey2026lostbackpropagationlmhead}.}
        \label{fig:prop_frozen_lm_head}
    \end{minipage}
    \hfill
    \begin{minipage}[t]{0.545\textwidth}
        \vspace{0pt}
        \centering
        \includegraphics[width=\linewidth]{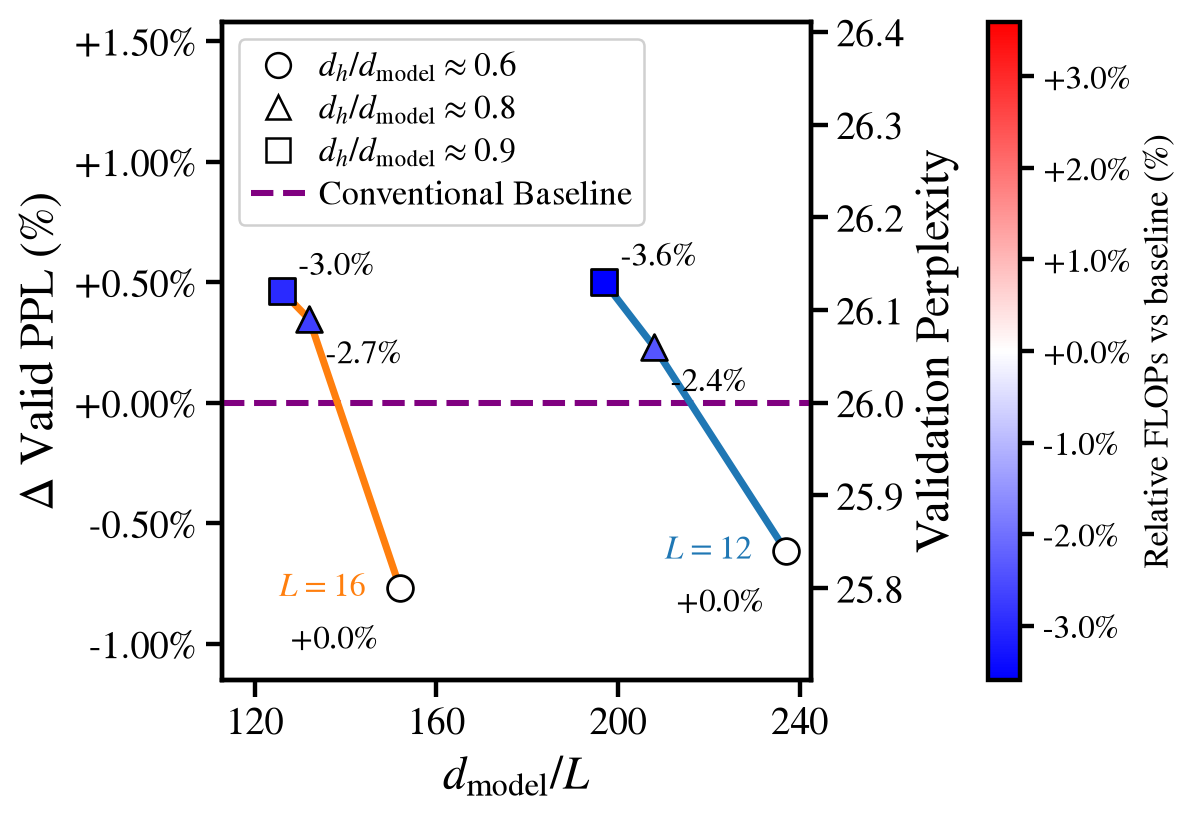}
        \captionsetup{hypcap=false}
        \captionof{figure}{\textbf{Width and depth of Hourglass Transformers at 906M.}
            Validation PPL delta versus $d_{\text{model}}/L$ for $L\in\{12,16\}$ and varying $d_h/d_{\text{model}}$ (marker shape). Lower $d_h/d_{\text{model}}\approx0.6$ gives the best PPL, while higher ratios provide larger FLOPs savings but move PPL above the baseline, revealing a quality--compute trade-off.}
        \label{fig:prop_ablation_curve_width_depth}
    \end{minipage}
\end{figure}

\autoref{fig:prop_frozen_lm_head} compares three configurations under the frozen-head setting.
Among the two conventional baselines alone, the variant with a wider $d_{\text{model}}$ ($d_{\text{model}}{=}1408$, $d_h{=}2112$, $d_h/d_{\text{model}}{=}1.5$, red) consistently outperforms the standard configuration ($d_{\text{model}}{=}1024$, $d_h{=}4096$, $d_h/d_{\text{model}}{=}4$, black), suggesting that a larger hidden dimension is beneficial in this setting.
The Hourglass model ($K{=}4$, $L{=}12$, $d_{\text{model}}{=}1952$, $d_h{=}1564$, $d_h/d_{\text{model}}{\approx}0.8$, blue) operates at the widest $d_{\text{model}}$ of the three and follows this trend.
This illustrates one advantage of the Hourglass architecture: it enables the use of a wider $d_{\text{model}}$ within the same parameter budget by compressing the FFN intermediate dimension.

% % \begin{figure}[t!]
% \begin{wrapfigure}{r}{0.53\textwidth}
%     \vspace{-7mm}
%     \centering
%     \includegraphics[width=\linewidth]{figures/frozen_loss_curves_v2.png}
%     \caption{\textbf{Frozen LM head Experiment at 403M.}
%         % Three models are compared: Conventional ($d_{\text{model}}{=}1024$, $d_h{=}4096$, black), Conventional with wider $d_{\text{model}}$ ($d_{\text{model}}{=}1408$, $d_h{=}2112$, red), and Hourglass ($K{=}4$, $L{=}12$, $d_{\text{model}}{=}1952$, $d_h{=}1564$, $d_h/d_{\text{model}}{\approx}0.8$, blue).
%         % The Hourglass curve uses $K{=}4$, $L{=}12$, $d_{\text{model}}{=}1952$, $d_h{=}1564$, and $d_h/d_{\text{model}}{\approx}0.8$.
%         The experiment probes the benefit of a larger $d_{\text{model}}$: the Hourglass model achieves the lowest loss, and the wider conventional variant outperforms the standard one, consistent with the gradient bottleneck hypothesis \citep{godey2026lostbackpropagationlmhead}.}
%     \label{fig:frozen_lm_head}
%     \vspace{-8mm}
% \end{wrapfigure}
% % \end{figure}

This ordering is consistent with the LM-head gradient bottleneck identified by \citet{godey2026lostbackpropagationlmhead}.
With a frozen head $W$, the backbone receives gradients $\nabla_H \mathcal{L}=\mathrm{diag}(\omega)(P-\tilde{N})W$, where $H$ denotes the pre-LM-head hidden states, $P$ the model's next-token probabilities, $\tilde{N}$ the empirical next-token distribution, and $\omega$ the per-position weights.
The learning signal is therefore compressed through the hidden dimension $D{=}d_{\text{model}}$; a wider $d_{\text{model}}$ alleviates this bottleneck.
Because the Hourglass architecture naturally widens $d_{\text{model}}$ (while keeping total parameters matched), it may benefit from both a more expressive backbone and a less constrained gradient pathway through the LM head.

% \begin{figure}[t!]
%     \centering
%     \begin{subfigure}[t]{0.45\textwidth}
%         \centering
%         \includegraphics[width=\textwidth]{figures/figure2_new1.png}
%         \caption{$d_h/d_{\text{model}}$ search at 403M}
%         \label{fig:ablation_curve}
%     \end{subfigure}
%     \hfill
%     \begin{subfigure}[t]{0.45\textwidth}
%         \centering
%         \includegraphics[width=\textwidth]{figures/906M_dmL_vs_validppl_flopsheat.png}
%         \caption{$d_{\text{model}}/L$ search at 906M}
%         \label{fig:ablation_curve_width_depth_900M_3B_8B}
%     \end{subfigure}
%     \caption{\textbf{Width and Depth of Hourglass Transformers.} (a) Validation perplexity across different $d_h/d_{\text{model}}$ ratios for Hourglass Transformers at 403M parameters. The $(K,L)=(4,12)$ curve (blue) remains below the conventional baseline across the sweep and is consistently lower than the $(K,L)=(4,24)$ curve (orange). (b) Validation PPL delta versus $d_{\text{model}}/L$ at 906M for Hourglass Transformers with $L\in\{12,16\}$ and varying $d_h/d_{\text{model}}$ (marker shape). Lower $d_h/d_{\text{model}}\approx0.6$ gives the best PPL, while higher ratios provide larger FLOPs savings but move PPL above the baseline, revealing a quality--compute trade-off.}
%     \vspace{-2mm}
% \end{figure}

\subsection{Width and Depth Landscape of Hourglass Transformers}
\label{sec:main_width_and_depth_landscape}
\vspace{-2mm}

In this section, we study the FFN intermediate ratio $d_h/d_\text{model}$ and the width-to-depth $d_\text{model}/L$ ratio of the model.
The central finding is that, within the Hourglass design space, models can remain competitive with the conventional baseline while using fewer Transformer layers.
By stacking deeper FFN sub-MLPs within each layer, the architecture frees parameters that can be allocated to widening $d_{\text{model}}$ rather than increasing the number of layers $L$.
We fix $K{=}4$ as a moderate-depth design choice that balances FFN compositional depth against residual-stream width under the matched parameter budget, rather than as the result of an exhaustive search over $K$ at every scale; Appendix~\ref{sec:ablation_vary_k} reports a small-scale ablation over $K$.
We establish this at the 403M scale (Appendix~\ref{app:fnn_intermediate_ratio_at_403M}) and then use a targeted sweep at 906M to identify reference configurations for scaling.

\autoref{fig:prop_ablation_curve_width_depth} shows the sweep at 906M parameters. Notably, the $(d_h/d_\text{model}, L) = (0.6, 12)$ and $(0.6, 16)$ configurations, using only 50\% and 67\% of the baseline's Transformer layers ($L=24$), exceed the baseline by 0.5\%, suggesting that the freed parameters are better spent on wider hidden representations.
We also observe a quality--compute trade-off: configurations with lower $d_h/d_\text{model}$ (circles, ${\approx}0.6$) reach the lowest validation PPL, while the largest savings (up to ${\sim}3\%$ fewer training FLOPs than the baseline) occur at higher ratios (squares, ${\approx}0.9$), at slightly worse PPL.
We therefore use this sweep to select an efficiency-oriented configuration for larger-scale validation.

\begin{table*}[t!]
    \centering
    \caption{\textbf{Configurations and quality--compute efficiency at 906M, 3B, and 8B.} Hourglass variants use hourglass-shaped FFNs with residual connections ($K$ stacked hourglass blocks), while conventional variants use standard narrow-wide-narrow FFNs ($K=1$). Training settings are summarized in \autoref{tab:exp_sum}. Parenthesized values in the Hourglass rows give the relative change versus the matched conventional baseline. The 64k decode column reports relative token-decode latency, where lower is faster. Training FLOPs are estimated by the script in Appendix~\ref{app:train_flops}; raw validation and downstream metrics are also provided in \autoref{tab:appendix_scale_results}.}
    \label{tab:baseline_and_hourglass_configs}
    \label{tab:paired_quality_efficiency}
    \setlength{\tabcolsep}{2.5pt}
    \renewcommand{\arraystretch}{1.08}
    \scriptsize
    \newcommand{\reldelta}[1]{{\tiny #1}}
    \resizebox{1\linewidth}{!}{
        \begin{tabular}{ll|l|llll|ll|lll}
            \toprule
            Size & Model
                 & \makecell[l]{Attn\ \ \ FFN\\Params (M)}
                 & $d_{\text{model}}$                             & $d_h$                                                  & $L$  & $K$
                 & \makecell[l]{Train FLOPs$\downarrow$}
                 & \makecell[l]{64k Decode\\Latency $\downarrow$}
                 & \makecell[l]{Val PPL $\downarrow$}
                 & \makecell[l]{Avg Downstream\\Acc. $\uparrow$}
                 & \makecell[l]{Avg Downstream\\PPL $\downarrow$}                                                                                                                                                                                                                                                     \\
            \midrule
            \multirow{2}{*}{906M}
                 & Conventional                                   & \makebox[2.2em][l]{227} \makebox[2.2em][l]{679}   & 1536 & 6144  & 24 & 1 & $3.34{\times}10^{16}$                        & $1.00{\times}$ & $26.00$                        & $0.503$                        & $1.524$                        \\
                 & Hourglass                                      & \makebox[2.2em][l]{184} \makebox[2.2em][l]{725}   & 2496 & 2016  & 12 & 4 & $3.26{\times}10^{16}$ \reldelta{($-2.40\%$)} & $0.52{\times}$ & $26.06$ \reldelta{($+0.23\%$)} & $0.505$ \reldelta{($+0.40\%$)} & $1.514$ \reldelta{($-0.66\%$)} \\
            \midrule
            \multirow{2}{*}{3B}
                 & Conventional                                   & \makebox[2.2em][l]{1057} \makebox[2.2em][l]{2114} & 3072 & 8192  & 28 & 1 & $1.03{\times}10^{17}$                        & $1.00{\times}$ & $16.80$                        & $0.626$                        & $1.211$                        \\
                 & Hourglass                                      & \makebox[2.2em][l]{817} \makebox[2.2em][l]{2350}  & 3696 & 2944  & 18 & 4 & $9.92{\times}10^{16}$ \reldelta{($-3.69\%$)} & $0.66{\times}$ & $16.93$ \reldelta{($+0.83\%$)} & $0.628$ \reldelta{($+0.32\%$)} & $1.188$ \reldelta{($-1.90\%$)} \\
            \midrule
            \multirow{2}{*}{8B}
                 & Conventional                                   & \makebox[2.2em][l]{2148} \makebox[2.2em][l]{5637} & 4096 & 14336 & 32 & 1 & $2.30{\times}10^{17}$                        & $1.00{\times}$ & $13.37$                        & $0.701$                        & $0.955$                        \\
                 & Hourglass                                      & \makebox[2.2em][l]{1799} \makebox[2.2em][l]{5988} & 4992 & 4544  & 22 & 4 & $2.25{\times}10^{17}$ \reldelta{($-2.17\%$)} & $0.70{\times}$ & $13.31$ \reldelta{($-0.45\%$)} & $0.706$ \reldelta{($+0.71\%$)} & $0.943$ \reldelta{($-1.26\%$)} \\
            \bottomrule
        \end{tabular}
    }
    \vspace{-2mm}
\end{table*}

\begin{figure*}[t!]
    \centering
    \begin{minipage}[t]{0.45\textwidth}
        \vspace{0pt}
        \centering
        \includegraphics[width=\linewidth]{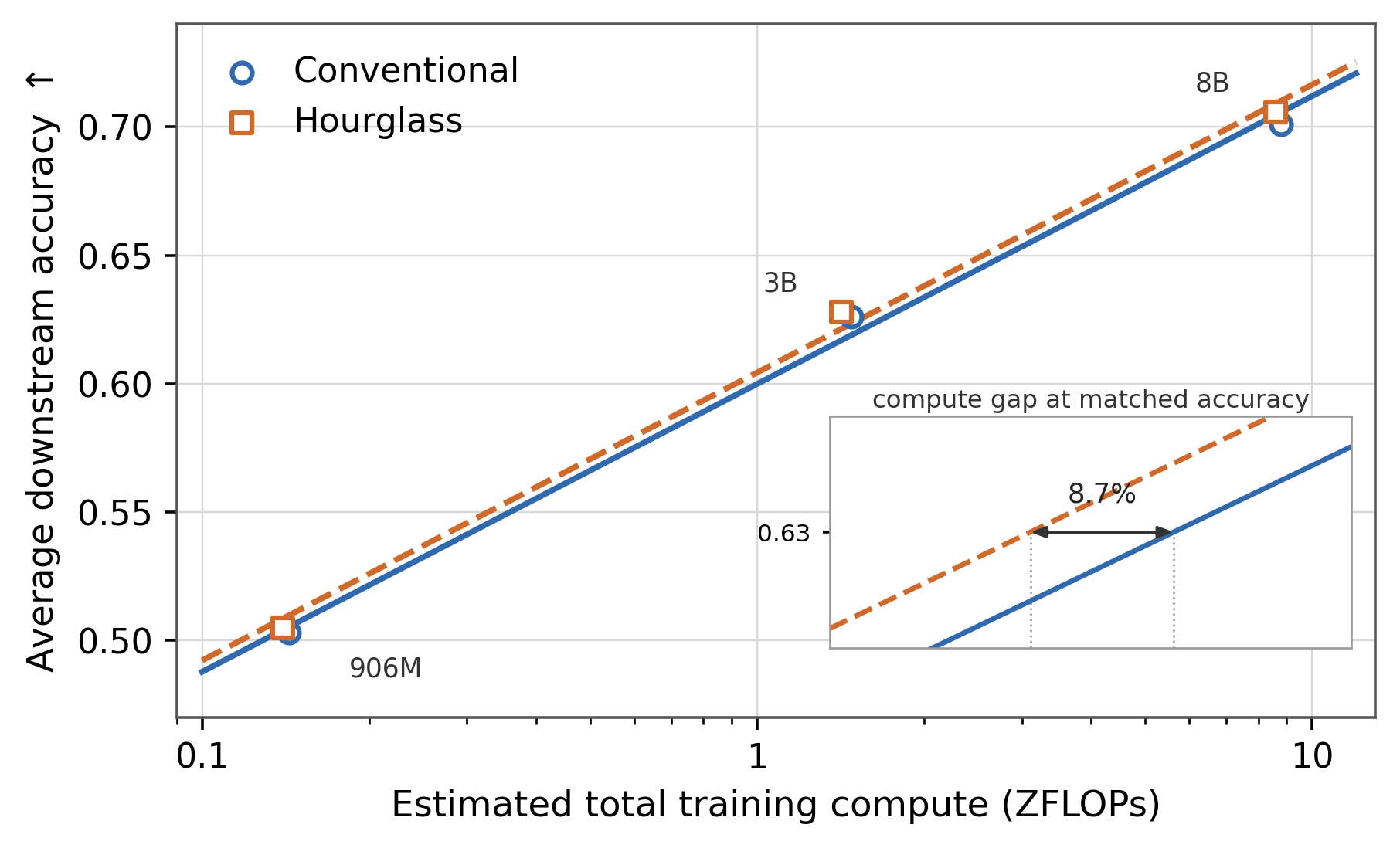}
        \captionsetup{hypcap=false}
        \captionof{figure}{\textbf{Average downstream accuracy versus estimated total training compute.}
            Lines are least-squares fits of $\mathrm{Acc}=a+b\log C$ with $R^2=0.996$ and $0.997$ for separate fits.
            Under the test regime, Hourglass reaches a matched average accuracy with $8.7\%$ less training compute.}
        \label{fig:compute_vs_downstream_acc}
    \end{minipage}
    \hfill
    \begin{minipage}[t]{0.52\textwidth}
        \vspace{10mm}
        \centering
        \setlength{\tabcolsep}{5pt}
        \renewcommand{\arraystretch}{1.0}
        \footnotesize
        \resizebox{\linewidth}{!}{
            \begin{tabular}{ll|ccccc|c}
                \toprule
                                                             & 8B Model     & 4k              & 8k             & 16k            & 32k            & 64k            & Avg.            \\
                \midrule
                \multirow{2}{*}{Proof-pile PPL $\downarrow$} & Conventional & $7.39$          & $6.81$         & $5.54$         & $5.05$         & $4.79$         & $5.92$          \\
                                                             & Hourglass    & $\mathbf{7.10}$ & $\mathbf{6.51}$ & $\mathbf{5.28}$ & $\mathbf{4.78}$ & $\mathbf{4.55}$ & $\mathbf{5.64}$ \\
                \midrule
                \multirow{2}{*}{RULER Acc. (\%) $\uparrow$}  & Conventional & $9.3$           & $5.8$          & $3.2$          & $2.5$          & $2.2$          & $4.6$           \\
                                                             & Hourglass    & $\mathbf{11.0}$ & $\mathbf{6.8}$ & $\mathbf{4.4}$ & $\mathbf{4.0}$ & $\mathbf{3.3}$ & $\mathbf{5.9}$  \\
                \bottomrule
            \end{tabular}
        }
        \vspace{1mm}
        \captionsetup{hypcap=false}
        \captionof{table}{\textbf{Long-context evaluation at 8B.} Both models receive an identical 400-step YaRN context extension~\citep{peng2023yarnefficientcontextwindow}; Hourglass uses $L=22$ attention layers versus $32$ for the matched conventional baseline. Perplexity is measured on Proof-pile and RULER accuracy is averaged over its 13 tasks. 
        Hourglass is better at every evaluated length on both metrics. 
        % The corresponding perplexities at 906M and 3B are reported in \autoref{tab:appendix_yarn_long_context}.
        }
        \label{tab:attn_yarn_long_context}
    \end{minipage}
    \vspace{-6mm}
\end{figure*}

\subsection{Scaling to 8B}
\label{sec:main_scales}
\vspace{-2mm}

Guided by the quality--compute frontier in the previous section, we scale an efficiency-oriented Hourglass architecture and compare it against the standard conventional baseline at 906M, 3B, and 8B parameters.
The selected configurations (\autoref{tab:baseline_and_hourglass_configs}) follow the design principles established above: deeper FFN stacking ($K{=}4$), fewer Transformer layers ($L{=}12/18/22$ at 906M/3B/8B versus $L{=}24/28/32$ for the baselines), wider $d_{\text{model}}$, and matched $d_{\text{attn}}$ with respective baselines.

\autoref{tab:paired_quality_efficiency} summarizes the paired quality--compute trade-off as raw values with relative deltas to the corresponding conventional baseline; the raw validation and downstream metrics are also reported in \autoref{tab:appendix_scale_results}.
% At these selected comparison points, Hourglass reduces total training FLOPs by $2.40\%$, $3.69\%$, and $2.17\%$ at 906M, 3B, and 8B, respectively.
Validation PPL remains close to the conventional baselines, with relative changes of $+0.23\%$, $+0.83\%$, and $-0.45\%$.
Downstream averages remain competitive: average accuracy changes by $+0.40\%$, $+0.32\%$, and $+0.71\%$, while average downstream PPL improves by $0.66\%$, $1.90\%$, and $1.26\%$.
Taking these pairs, we fit quality on total training compute $C=(\text{FLOPs/token})\times\text{tokens}$ to understand the training efficient improvment between the two architectures at matched performance (\autoref{fig:compute_vs_downstream_acc}), revealing that Hourglass reaches a given average accuracy with $8.7\%$ less training compute.
Overall, the scaled Hourglass architecture preserves competitive validation and downstream quality while using modestly fewer training FLOPs across 906M, 3B, and 8B.
Since FFN shape, residual width, layer count, and attention width are varied jointly, these results support the combined Hourglass design as a whole.

\subsection{Long-Context Language Modeling}
\label{sec:long_context}
\vspace{-2mm}

Hourglass reduces the number of Transformer layers, which gives latency gains but also removes attention applications that might otherwise support long-range mixing.
We evaluate the 8B conventional/Hourglass pairs at \autoref{tab:baseline_and_hourglass_configs} after identical 400-step YaRN finetuning for context extension up to 64k tokens, using Proof-pile following \citet{peng2023yarnefficientcontextwindow} and RULER~\citep{hsieh2024ruler}, whose 13 tasks probe retrieval, multi-key and multi-query needle search, aggregation, variable tracking, and question answering.
% Because perplexity alone does not establish that a model can use its context, we additionally evaluate the 8B pair on RULER~\citep{hsieh2024ruler}, whose 13 tasks probe retrieval, multi-key and multi-query needle search, aggregation, variable tracking, and question answering.
% \autoref{tab:attn_yarn_long_context} reports both metrics for the 8B pair, and \autoref{tab:appendix_yarn_long_context} reports the perplexities at all three scales.
Notably, Hourglass obtains lower PPL and higher RULER accuracy at every evaluated context length.
% At 64k, PPL improves from 9.59 to 9.38 at 906M, from 6.88 to 6.76 at 3B, and from 4.79 to 4.55 at 8B; the largest relative improvements occur at intermediate lengths, reaching $4.8\%$ at 906M, $3.3\%$ at 3B, and $5.4\%$ at 8B.
% The functional evaluation moves in the same direction: at 8B, RULER accuracy improves at every length, from $4.6\%$ to $5.9\%$ on average, a gain of $1.0$--$1.7$ percentage points per length.
% We read this as complementary paired evidence rather than a strong long-context claim, since the absolute RULER scores remain low at this training budget for both architectures.
One possible explanation is that the wider residual embedding space in Hourglass can preserve token information better across long contexts, and the fewer attention layers reduce accumulated transformation/noise during very long-context decoding.

\subsection{Token Decode Efficiency}
\label{sec:latency}
\vspace{-2mm}

\begin{wrapfigure}{r}{0.47\textwidth}
    \vspace{-6mm}
    \centering
    \includegraphics[width=\linewidth]{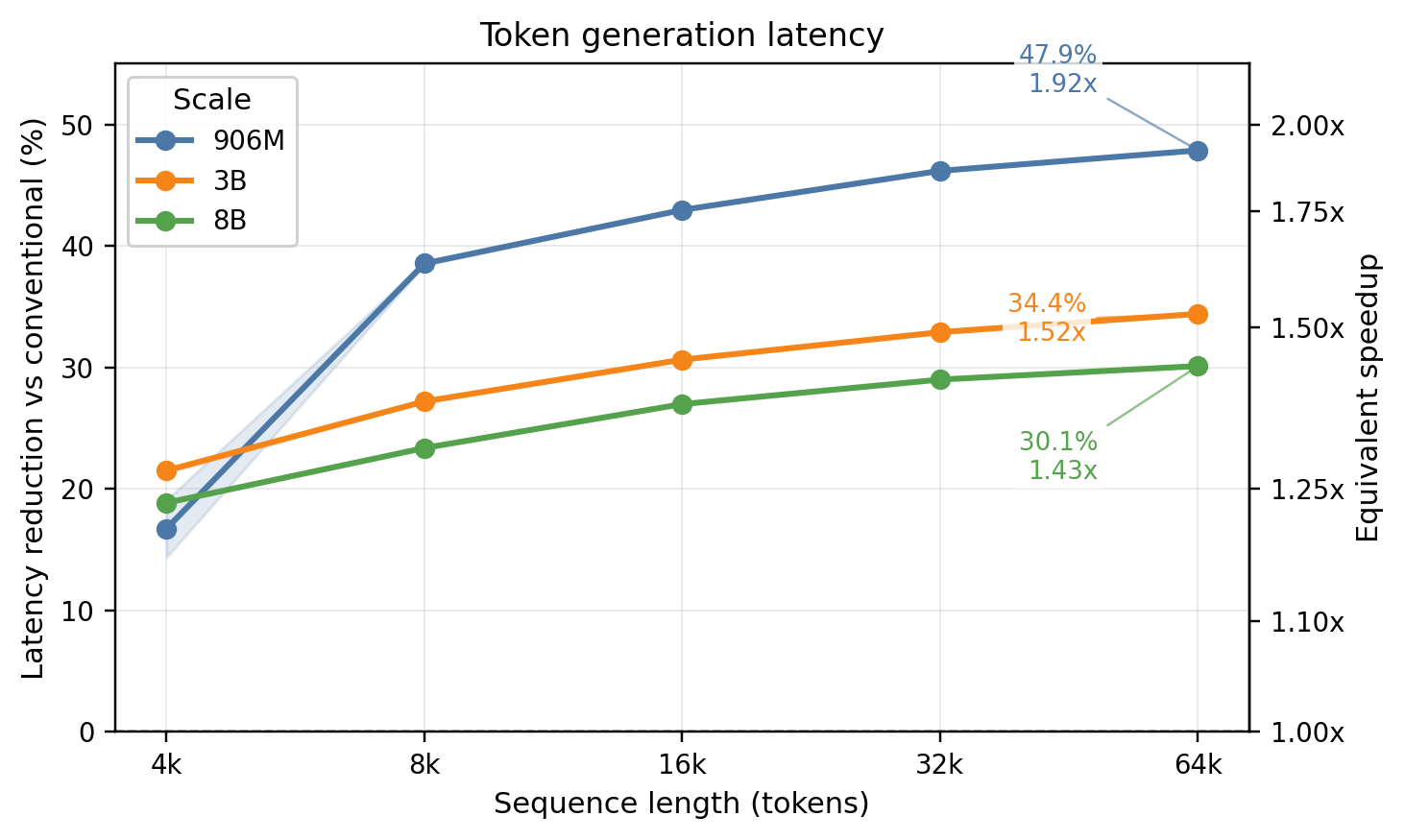}
    \caption{\textbf{Token decode latency.} Relative latency is normalized to the conventional baseline; endpoint labels report the 64k-token values.}
    \label{fig:latency_comparison}
    \vspace{-6mm}
\end{wrapfigure}

A practical benefit of the Hourglass architecture using fewer Transformer layers than the conventional baseline at the same parameter budget is that the reduced $L$ directly translates to lower KV cache size and thus improved decode latency.
% A practical benefit of the Hourglass architecture is that it uses fewer Transformer layers than the conventional baseline at the same parameter budget, where reducing $L$ directly translates to lower KV cache size and thus improved decode latency.
Figure~\ref{fig:latency_comparison} reports relative token decode latency across sequence lengths from 4k to 64k tokens at 906M, 3B, and 8B.
As context length increases, the speedup aymptotically approaches the $L_{\text{conv}}/L_{\text{hg}}$ ratio, which is $2\times$, $1.56\times$, and $1.45\times$ at 906M, 3B, and 8B, respectively. We additionaly evaluate with different sequence mixers (GQA and MLA) in \autoref{tab:mixer_decode_latency} for 906M model up to 128k context length and observe the same speedup trend.

\vspace{-2mm}
\section{Discussion and Future Work}
\label{sec:discussion}
\vspace{-2mm}

Our investigation revisits the long-standing shape convention of dense Transformer language models, with particular focus on the narrow-wide-narrow FFN.
Our evaluation results suggest that the conventional FFN expansion is not essential for maintaining competitive language-model quality: residual hourglass bottlenecks can provide an alternative token-wise transformation path, while freeing capacity for wider residual representations and fewer Transformer layers.
Decoupling attention width from residual-stream width makes this reallocation practical, allowing sequence mixing to remain close to the baseline attention dimension even as the residual stream widens.

\vspace{-2mm}

\paragraph{Limitations.}
While these results are promising, specific limitations remain.
First, our comprehensive parameter search was constrained to smaller scales, with the selected recipe validated up to 8B parameters trained with 160B tokens. Validation at larger parameter size and training budget to trillion tokens for short and long contexts could further confirm the generality of the hourglass design.
% Consequently, robust scaling beyond 8B remains an open empirical question and may require additional architecture tuning.
Second, while our controlled ablations isolate the Hourglass FFN mechanism at smaller scales, the 906M, 3B, and 8B experiments validate the complete architecture recipe, where FFN shape, residual width, layer count, and attention projection width are selected jointly.
Thus, the main large-scale conclusion is an end-to-end architecture-efficiency result. Full factor decomposition across every design axis with detailed analysis remains a future direction.
Third, all trained models use multi-head attention as the sequence mixer.
Appendix~\ref{app:mixer_decode_latency} measures decode cost with Group-Query Attention \citep{jiang2023mistral7b} and Multi-head Latent Attention \citep{deepseekai2025deepseekv3technicalreport} substituted into both arms, and finds the KV-cache and latency advantages preserved, but those measurements concern inference cost only.
How these mixers interact with Hourglass FFNs during training, and whether they shift the attention-FFN ratio landscape, remains unexplored.

\vspace{-2mm}

\paragraph{Future Work.}
Continued scaling of model dimensions ($d_{\text{model}}$ and $K$) is a promising avenue, as similar manifold-constrained architectures have shown benefits from expanded width \cite{xie2025mhc}.
At the same time, deeper hourglass FFNs may require careful control of signal propagation and residual noise, especially in regimes affected by the "curse of depth" \cite{sun2025curse}.
Broader long-context evaluation is another natural next step, including retrieval-heavy tasks, instruction-following settings, reasoning-oriented benchmarks, and context windows beyond 64k.
We view Hourglass Transformers as a practical step toward expanding the design space of efficient dense language models beyond the historical narrow-wide-narrow FFN convention.

\bibliographystyle{plainnat}
\bibliography{neurips_2026}

@misc{jiang2023mistral7b,
  title         = {Mistral 7B},
  author        = {Albert Q. Jiang and Alexandre Sablayrolles and Arthur Mensch and Chris Bamford and Devendra Singh Chaplot and Diego de las Casas and Florian Bressand and Gianna Lengyel and Guillaume Lample and Lucile Saulnier and Lélio Renard Lavaud and Marie-Anne Lachaux and Pierre Stock and Teven Le Scao and Thibaut Lavril and Thomas Wang and Timothée Lacroix and William El Sayed},
  year          = {2023},
  eprint        = {2310.06825},
  archiveprefix = {arXiv},
  primaryclass  = {cs.CL},
  url           = {https://arxiv.org/abs/2310.06825}
}

@misc{deepseekai2025deepseekv3technicalreport,
  title         = {DeepSeek-V3 Technical Report},
  author        = {DeepSeek-AI and Aixin Liu and Bei Feng and Bing Xue and Bingxuan Wang and Bochao Wu and Chengda Lu and Chenggang Zhao and Chengqi Deng and Chenyu Zhang and Chong Ruan and Damai Dai and Daya Guo and Dejian Yang and Deli Chen and Dongjie Ji and Erhang Li and Fangyun Lin and Fucong Dai and Fuli Luo and Guangbo Hao and Guanting Chen and Guowei Li and H. Zhang and Han Bao and Hanwei Xu and Haocheng Wang and Haowei Zhang and Honghui Ding and Huajian Xin and Huazuo Gao and Hui Li and Hui Qu and J. L. Cai and Jian Liang and Jianzhong Guo and Jiaqi Ni and Jiashi Li and Jiawei Wang and Jin Chen and Jingchang Chen and Jingyang Yuan and Junjie Qiu and Junlong Li and Junxiao Song and Kai Dong and Kai Hu and Kaige Gao and Kang Guan and Kexin Huang and Kuai Yu and Lean Wang and Lecong Zhang and Lei Xu and Leyi Xia and Liang Zhao and Litong Wang and Liyue Zhang and Meng Li and Miaojun Wang and Mingchuan Zhang and Minghua Zhang and Minghui Tang and Mingming Li and Ning Tian and Panpan Huang and Peiyi Wang and Peng Zhang and Qiancheng Wang and Qihao Zhu and Qinyu Chen and Qiushi Du and R. J. Chen and R. L. Jin and Ruiqi Ge and Ruisong Zhang and Ruizhe Pan and Runji Wang and Runxin Xu and Ruoyu Zhang and Ruyi Chen and S. S. Li and Shanghao Lu and Shangyan Zhou and Shanhuang Chen and Shaoqing Wu and Shengfeng Ye and Shengfeng Ye and Shirong Ma and Shiyu Wang and Shuang Zhou and Shuiping Yu and Shunfeng Zhou and Shuting Pan and T. Wang and Tao Yun and Tian Pei and Tianyu Sun and W. L. Xiao and Wangding Zeng and Wanjia Zhao and Wei An and Wen Liu and Wenfeng Liang and Wenjun Gao and Wenqin Yu and Wentao Zhang and X. Q. Li and Xiangyue Jin and Xianzu Wang and Xiao Bi and Xiaodong Liu and Xiaohan Wang and Xiaojin Shen and Xiaokang Chen and Xiaokang Zhang and Xiaosha Chen and Xiaotao Nie and Xiaowen Sun and Xiaoxiang Wang and Xin Cheng and Xin Liu and Xin Xie and Xingchao Liu and Xingkai Yu and Xinnan Song and Xinxia Shan and Xinyi Zhou and Xinyu Yang and Xinyuan Li and Xuecheng Su and Xuheng Lin and Y. K. Li and Y. Q. Wang and Y. X. Wei and Y. X. Zhu and Yang Zhang and Yanhong Xu and Yanhong Xu and Yanping Huang and Yao Li and Yao Zhao and Yaofeng Sun and Yaohui Li and Yaohui Wang and Yi Yu and Yi Zheng and Yichao Zhang and Yifan Shi and Yiliang Xiong and Ying He and Ying Tang and Yishi Piao and Yisong Wang and Yixuan Tan and Yiyang Ma and Yiyuan Liu and Yongqiang Guo and Yu Wu and Yuan Ou and Yuchen Zhu and Yuduan Wang and Yue Gong and Yuheng Zou and Yujia He and Yukun Zha and Yunfan Xiong and Yunxian Ma and Yuting Yan and Yuxiang Luo and Yuxiang You and Yuxuan Liu and Yuyang Zhou and Z. F. Wu and Z. Z. Ren and Zehui Ren and Zhangli Sha and Zhe Fu and Zhean Xu and Zhen Huang and Zhen Zhang and Zhenda Xie and Zhengyan Zhang and Zhewen Hao and Zhibin Gou and Zhicheng Ma and Zhigang Yan and Zhihong Shao and Zhipeng Xu and Zhiyu Wu and Zhongyu Zhang and Zhuoshu Li and Zihui Gu and Zijia Zhu and Zijun Liu and Zilin Li and Ziwei Xie and Ziyang Song and Ziyi Gao and Zizheng Pan},
  year          = {2025},
  eprint        = {2412.19437},
  archiveprefix = {arXiv},
  primaryclass  = {cs.CL},
  url           = {https://arxiv.org/abs/2412.19437}
}

@misc{olmo20252olmo2furious,
  title         = {2 OLMo 2 Furious},
  author        = {Team OLMo and Pete Walsh and Luca Soldaini and Dirk Groeneveld and Kyle Lo and Shane Arora and Akshita Bhagia and Yuling Gu and Shengyi Huang and Matt Jordan and Nathan Lambert and Dustin Schwenk and Oyvind Tafjord and Taira Anderson and David Atkinson and Faeze Brahman and Christopher Clark and Pradeep Dasigi and Nouha Dziri and Allyson Ettinger and Michal Guerquin and David Heineman and Hamish Ivison and Pang Wei Koh and Jiacheng Liu and Saumya Malik and William Merrill and Lester James V. Miranda and Jacob Morrison and Tyler Murray and Crystal Nam and Jake Poznanski and Valentina Pyatkin and Aman Rangapur and Michael Schmitz and Sam Skjonsberg and David Wadden and Christopher Wilhelm and Michael Wilson and Luke Zettlemoyer and Ali Farhadi and Noah A. Smith and Hannaneh Hajishirzi},
  year          = {2025},
  eprint        = {2501.00656},
  archiveprefix = {arXiv},
  primaryclass  = {cs.CL},
  url           = {https://arxiv.org/abs/2501.00656}
}

@misc{peng2023yarnefficientcontextwindow,
  title         = {YaRN: Efficient Context Window Extension of Large Language Models},
  author        = {Bowen Peng and Jeffrey Quesnelle and Honglu Fan and Enrico Shippole},
  year          = {2023},
  eprint        = {2309.00071},
  archiveprefix = {arXiv},
  primaryclass  = {cs.CL},
  url           = {https://arxiv.org/abs/2309.00071}
}

@inproceedings{vaswani2017attentionneed,
  author    = {Vaswani, Ashish and Shazeer, Noam and Parmar, Niki and Uszkoreit, Jakob and Jones, Llion and Gomez, Aidan N and Kaiser, \L ukasz and Polosukhin, Illia},
  booktitle = {Advances in Neural Information Processing Systems},
  editor    = {I. Guyon and U. Von Luxburg and S. Bengio and H. Wallach and R. Fergus and S. Vishwanathan and R. Garnett},
  pages     = {},
  publisher = {Curran Associates, Inc.},
  title     = {Attention is All you Need},
  url       = {https://proceedings.neurips.cc/paper_files/paper/2017/file/3f5ee243547dee91fbd053c1c4a845aa-Paper.pdf},
  volume    = {30},
  year      = {2017}
}

@misc{chen2025rethinkingshapeconventionmlp,
  title         = {Rethinking the shape convention of an MLP},
  author        = {Meng-Hsi Chen and Yu-Ang Lee and Feng-Ting Liao and Da-shan Shiu},
  year          = {2025},
  eprint        = {2510.01796},
  archiveprefix = {arXiv},
  primaryclass  = {cs.LG},
  url           = {https://arxiv.org/abs/2510.01796}
}

@inproceedings{10.1007/978-3-032-09645-6_10,
  author    = {Joyce, Johnny
               and Verschelde, Jan},
  editor    = {Boulier, Fran{\c{c}}ois
               and Mou, Chenqi
               and Sadykov, Timur M.
               and Vorozhtsov, Evgenii V.},
  title     = {Computing Linear Regions in Neural Networks with Skip Connections},
  booktitle = {Computer Algebra in Scientific Computing},
  year      = {2026},
  publisher = {Springer Nature Switzerland},
  address   = {Cham},
  pages     = {175--194},
  isbn      = {978-3-032-09645-6}
}

@inproceedings{NEURIPS2018_03bfc1d4,
  author    = {Lin, Hongzhou and Jegelka, Stefanie},
  booktitle = {Advances in Neural Information Processing Systems},
  editor    = {S. Bengio and H. Wallach and H. Larochelle and K. Grauman and N. Cesa-Bianchi and R. Garnett},
  pages     = {},
  publisher = {Curran Associates, Inc.},
  title     = {ResNet with one-neuron hidden layers is a Universal Approximator},
  url       = {https://proceedings.neurips.cc/paper_files/paper/2018/file/03bfc1d4783966c69cc6aef8247e0103-Paper.pdf},
  volume    = {31},
  year      = {2018}
}

@inproceedings{pmlr-v235-liu24am,
  title     = {Characterizing {R}es{N}et’s Universal Approximation Capability},
  author    = {Liu, Chenghao and Liang, Enming and Chen, Minghua},
  booktitle = {Proceedings of the 41st International Conference on Machine Learning},
  pages     = {31477--31515},
  year      = {2024},
  editor    = {Salakhutdinov, Ruslan and Kolter, Zico and Heller, Katherine and Weller, Adrian and Oliver, Nuria and Scarlett, Jonathan and Berkenkamp, Felix},
  volume    = {235},
  series    = {Proceedings of Machine Learning Research},
  month     = {21--27 Jul},
  publisher = {PMLR},
  url       = {https://proceedings.mlr.press/v235/liu24am.html}
}

@inproceedings{ICML-2019-HaninR,
  author    = {Boris Hanin and David Rolnick},
  booktitle = {{Proceedings of the 36th International Conference on Machine Learning}},
  ee        = {http://proceedings.mlr.press/v97/hanin19a.html},
  pages     = {2596--2604},
  publisher = {{PMLR}},
  title     = {{Complexity of Linear Regions in Deep Networks}},
  year      = 2019
}

@article{cover1965covertheorm,
  author   = {Cover, Thomas M.},
  journal  = {IEEE Transactions on Electronic Computers},
  title    = {Geometrical and Statistical Properties of Systems of Linear Inequalities with Applications in Pattern Recognition},
  year     = {1965},
  volume   = {EC-14},
  number   = {3},
  pages    = {326-334},
  doi      = {10.1109/PGEC.1965.264137}
}

@inproceedings{Rahimi2007,
  author    = {Rahimi, Ali and Recht, Benjamin},
  title     = {Random Features for Large-Scale Kernel Machines},
  booktitle = {Proceedings of the 20th International Conference on Neural Information Processing Systems},
  series    = {NIPS'07},
  year      = {2007},
  location  = {Vancouver, British Columbia, Canada},
  pages     = {1177–1184},
  numpages  = {8},
  publisher = {Curran Associates Inc.},
  address   = {Red Hook, NY, USA}
}

@inproceedings{10.1007/978-3-030-58580-8_40,
  author    = {Zhou, Daquan and Hou, Qibin and Chen, Yunpeng and Feng, Jiashi and Yan, Shuicheng},
  title     = {Rethinking Bottleneck Structure for Efficient Mobile Network Design},
  year      = {2020},
  isbn      = {978-3-030-58579-2},
  publisher = {Springer-Verlag},
  address   = {Berlin, Heidelberg},
  url       = {https://doi.org/10.1007/978-3-030-58580-8_40},
  doi       = {10.1007/978-3-030-58580-8_40},
  booktitle = {Computer Vision – ECCV 2020: 16th European Conference, Glasgow, UK, August 23–28, 2020, Proceedings, Part III},
  pages     = {680–697},
  numpages  = {18},
  location  = {Glasgow, United Kingdom}
}

@inproceedings{shazeer2017outrageously,
  author    = {Shazeer, Noam and Mirhoseini, Azalia and Maziarz, Krzysztof and Davis, Andy and Le, Quoc and Hinton, Geoffrey and Dean, Jeff},
  title     = {Outrageously Large Neural Networks: The Sparsely-Gated Mixture-of-Experts Layer},
  booktitle = {International Conference on Learning Representations (ICLR)},
  year      = {2017},
  url       = {https://arxiv.org/abs/1701.06538}
}

@inproceedings{hu2022lora,
  title     = {{LoRA}: Low-Rank Adaptation of Large Language Models},
  author    = {Hu, Edward J. and Shen, Yelong and Wallis, Phillip and Allen-Zhu, Zeyuan and Li, Yuanzhi and Wang, Shean and Wang, Lu and Chen, Weizhu},
  booktitle = {International Conference on Learning Representations (ICLR)},
  year      = {2022},
  url       = {https://openreview.net/forum?id=nZe72R8yS0}
}

@article{touvron2023llama,
  title   = {Llama: Open and Efficient Foundation Language Models},
  author  = {Touvron, Hugo and Lavril, Thibaut and Izacard, Gautier and Martinet, Xavier and Lachaux, Marie-Anne and Lacroix, Timoth{\'e}e and Rozi{\`e}re, Baptiste and Goyal, Naman and Hambro, Eric and Azhar, Faisal and Rodriguez, Aurelien and Joulin, Armand and Grave, Edouard and Lample, Guillaume},
  journal = {arXiv preprint arXiv:2302.13971},
  year    = {2023},
  url     = {https://arxiv.org/abs/2302.13971}
}

@misc{qwen2025qwen25technicalreport,
  title         = {Qwen2.5 Technical Report},
  author        = {Qwen and : and An Yang and Baosong Yang and Beichen Zhang and Binyuan Hui and Bo Zheng and Bowen Yu and Chengyuan Li and Dayiheng Liu and Fei Huang and Haoran Wei and Huan Lin and Jian Yang and Jianhong Tu and Jianwei Zhang and Jianxin Yang and Jiaxi Yang and Jingren Zhou and Junyang Lin and Kai Dang and Keming Lu and Keqin Bao and Kexin Yang and Le Yu and Mei Li and Mingfeng Xue and Pei Zhang and Qin Zhu and Rui Men and Runji Lin and Tianhao Li and Tianyi Tang and Tingyu Xia and Xingzhang Ren and Xuancheng Ren and Yang Fan and Yang Su and Yichang Zhang and Yu Wan and Yuqiong Liu and Zeyu Cui and Zhenru Zhang and Zihan Qiu},
  year          = {2025},
  eprint        = {2412.15115},
  archiveprefix = {arXiv},
  primaryclass  = {cs.CL},
  url           = {https://arxiv.org/abs/2412.15115}
}

@misc{gemmateam2025gemma3technicalreport,
  title         = {Gemma 3 Technical Report},
  author        = {Gemma Team and Aishwarya Kamath and Johan Ferret and Shreya Pathak and Nino Vieillard and Ramona Merhej and Sarah Perrin and Tatiana Matejovicova and Alexandre Ramé and Morgane Rivière and Louis Rouillard and Thomas Mesnard and Geoffrey Cideron and Jean-bastien Grill and Sabela Ramos and Edouard Yvinec and Michelle Casbon and Etienne Pot and Ivo Penchev and Gaël Liu and Francesco Visin and Kathleen Kenealy and Lucas Beyer and Xiaohai Zhai and Anton Tsitsulin and Robert Busa-Fekete and Alex Feng and Noveen Sachdeva and Benjamin Coleman and Yi Gao and Basil Mustafa and Iain Barr and Emilio Parisotto and David Tian and Matan Eyal and Colin Cherry and Jan-Thorsten Peter and Danila Sinopalnikov and Surya Bhupatiraju and Rishabh Agarwal and Mehran Kazemi and Dan Malkin and Ravin Kumar and David Vilar and Idan Brusilovsky and Jiaming Luo and Andreas Steiner and Abe Friesen and Abhanshu Sharma and Abheesht Sharma and Adi Mayrav Gilady and Adrian Goedeckemeyer and Alaa Saade and Alex Feng and Alexander Kolesnikov and Alexei Bendebury and Alvin Abdagic and Amit Vadi and András György and André Susano Pinto and Anil Das and Ankur Bapna and Antoine Miech and Antoine Yang and Antonia Paterson and Ashish Shenoy and Ayan Chakrabarti and Bilal Piot and Bo Wu and Bobak Shahriari and Bryce Petrini and Charlie Chen and Charline Le Lan and Christopher A. Choquette-Choo and CJ Carey and Cormac Brick and Daniel Deutsch and Danielle Eisenbud and Dee Cattle and Derek Cheng and Dimitris Paparas and Divyashree Shivakumar Sreepathihalli and Doug Reid and Dustin Tran and Dustin Zelle and Eric Noland and Erwin Huizenga and Eugene Kharitonov and Frederick Liu and Gagik Amirkhanyan and Glenn Cameron and Hadi Hashemi and Hanna Klimczak-Plucińska and Harman Singh and Harsh Mehta and Harshal Tushar Lehri and Hussein Hazimeh and Ian Ballantyne and Idan Szpektor and Ivan Nardini and Jean Pouget-Abadie and Jetha Chan and Joe Stanton and John Wieting and Jonathan Lai and Jordi Orbay and Joseph Fernandez and Josh Newlan and Ju-yeong Ji and Jyotinder Singh and Kat Black and Kathy Yu and Kevin Hui and Kiran Vodrahalli and Klaus Greff and Linhai Qiu and Marcella Valentine and Marina Coelho and Marvin Ritter and Matt Hoffman and Matthew Watson and Mayank Chaturvedi and Michael Moynihan and Min Ma and Nabila Babar and Natasha Noy and Nathan Byrd and Nick Roy and Nikola Momchev and Nilay Chauhan and Noveen Sachdeva and Oskar Bunyan and Pankil Botarda and Paul Caron and Paul Kishan Rubenstein and Phil Culliton and Philipp Schmid and Pier Giuseppe Sessa and Pingmei Xu and Piotr Stanczyk and Pouya Tafti and Rakesh Shivanna and Renjie Wu and Renke Pan and Reza Rokni and Rob Willoughby and Rohith Vallu and Ryan Mullins and Sammy Jerome and Sara Smoot and Sertan Girgin and Shariq Iqbal and Shashir Reddy and Shruti Sheth and Siim Põder and Sijal Bhatnagar and Sindhu Raghuram Panyam and Sivan Eiger and Susan Zhang and Tianqi Liu and Trevor Yacovone and Tyler Liechty and Uday Kalra and Utku Evci and Vedant Misra and Vincent Roseberry and Vlad Feinberg and Vlad Kolesnikov and Woohyun Han and Woosuk Kwon and Xi Chen and Yinlam Chow and Yuvein Zhu and Zichuan Wei and Zoltan Egyed and Victor Cotruta and Minh Giang and Phoebe Kirk and Anand Rao and Kat Black and Nabila Babar and Jessica Lo and Erica Moreira and Luiz Gustavo Martins and Omar Sanseviero and Lucas Gonzalez and Zach Gleicher and Tris Warkentin and Vahab Mirrokni and Evan Senter and Eli Collins and Joelle Barral and Zoubin Ghahramani and Raia Hadsell and Yossi Matias and D. Sculley and Slav Petrov and Noah Fiedel and Noam Shazeer and Oriol Vinyals and Jeff Dean and Demis Hassabis and Koray Kavukcuoglu and Clement Farabet and Elena Buchatskaya and Jean-Baptiste Alayrac and Rohan Anil and Dmitry and Lepikhin and Sebastian Borgeaud and Olivier Bachem and Armand Joulin and Alek Andreev and Cassidy Hardin and Robert Dadashi and Léonard Hussenot},
  year          = {2025},
  eprint        = {2503.19786},
  archiveprefix = {arXiv},
  primaryclass  = {cs.CL},
  url           = {https://arxiv.org/abs/2503.19786}
}

@inproceedings{BMVC2016_87,
  title     = {Wide Residual Networks},
  author    = {Sergey Zagoruyko and Nikos Komodakis},
  year      = {2016},
  month     = {September},
  pages     = {87.1-87.12},
  articleno = {87},
  numpages  = {12},
  booktitle = {Proceedings of the British Machine Vision Conference (BMVC)},
  publisher = {BMVA Press},
  editor    = {Richard C. Wilson, Edwin R. Hancock and William A. P. Smith},
  doi       = {10.5244/C.30.87},
  isbn      = {1-901725-59-6},
  url       = {https://dx.doi.org/10.5244/C.30.87}
}

@inproceedings{10.1007/978-3-319-24574-4_28,
  author    = {Ronneberger, Olaf
               and Fischer, Philipp
               and Brox, Thomas},
  editor    = {Navab, Nassir
               and Hornegger, Joachim
               and Wells, William M.
               and Frangi, Alejandro F.},
  title     = {U-Net: Convolutional Networks for Biomedical Image Segmentation},
  booktitle = {Medical Image Computing and Computer-Assisted Intervention -- MICCAI 2015},
  year      = {2015},
  publisher = {Springer International Publishing},
  address   = {Cham},
  pages     = {234--241},
  isbn      = {978-3-319-24574-4}
}

@inproceedings{10.5555/3692070.3693369,
  author    = {Liu, Shih-Yang and Wang, Chien-Yi and Yin, Hongxu and Molchanov, Pavlo and Wang, Yu-Chiang Frank and Cheng, Kwang-Ting and Chen, Min-Hung},
  title     = {DoRA: weight-decomposed low-rank adaptation},
  year      = {2024},
  publisher = {JMLR.org},
  booktitle = {Proceedings of the 41st International Conference on Machine Learning},
  articleno = {1299},
  numpages  = {22},
  location  = {Vienna, Austria},
  series    = {ICML'24}
}

@misc{mu2025comprehensivesurveymixtureofexpertsalgorithms,
  title         = {A Comprehensive Survey of Mixture-of-Experts: Algorithms, Theory, and Applications},
  author        = {Siyuan Mu and Sen Lin},
  year          = {2025},
  eprint        = {2503.07137},
  archiveprefix = {arXiv},
  primaryclass  = {cs.LG},
  url           = {https://arxiv.org/abs/2503.07137}
}

@misc{shazeer2020gluvariantsimprovetransformer,
  title         = {GLU Variants Improve Transformer},
  author        = {Noam Shazeer},
  year          = {2020},
  eprint        = {2002.05202},
  archiveprefix = {arXiv},
  primaryclass  = {cs.LG},
  url           = {https://arxiv.org/abs/2002.05202}
}

@misc{xiong2020on,
  title  = {On Layer Normalization in the Transformer Architecture},
  author = {Ruibin Xiong and Yunchang Yang and Di He and Kai Zheng and Shuxin Zheng and Huishuai Zhang and Yanyan Lan and Liwei Wang and Tie-Yan Liu},
  year   = {2020},
  url    = {https://openreview.net/forum?id=B1x8anVFPr}
}

@inproceedings{jiang2023prermsnorm,
  title     = {Pre-{RMSN}orm and Pre-{CRMSN}orm Transformers: Equivalent and Efficient Pre-{LN} Transformers},
  author    = {Zixuan Jiang and Jiaqi Gu and Hanqing Zhu and David Z. Pan},
  booktitle = {Thirty-seventh Conference on Neural Information Processing Systems},
  year      = {2023},
  url       = {https://openreview.net/forum?id=z06npyCwDq}
}

@inproceedings{li-etal-2022-ode,
  title     = {{ODE} Transformer: An Ordinary Differential Equation-Inspired Model for Sequence Generation},
  author    = {Li, Bei  and
               Du, Quan  and
               Zhou, Tao  and
               Jing, Yi  and
               Zhou, Shuhan  and
               Zeng, Xin  and
               Xiao, Tong  and
               Zhu, JingBo  and
               Liu, Xuebo  and
               Zhang, Min},
  editor    = {Muresan, Smaranda  and
               Nakov, Preslav  and
               Villavicencio, Aline},
  booktitle = {Proceedings of the 60th Annual Meeting of the Association for Computational Linguistics (Volume 1: Long Papers)},
  month     = may,
  year      = {2022},
  address   = {Dublin, Ireland},
  publisher = {Association for Computational Linguistics},
  url       = {https://aclanthology.org/2022.acl-long.571/},
  doi       = {10.18653/v1/2022.acl-long.571},
  pages     = {8335--8351}
}

@inproceedings{Lan2020ALBERT,
  title     = {ALBERT: A Lite BERT for Self-supervised Learning of Language Representations},
  author    = {Zhenzhong Lan and Mingda Chen and Sebastian Goodman and Kevin Gimpel and Piyush Sharma and Radu Soricut},
  booktitle = {International Conference on Learning Representations},
  year      = {2020},
  url       = {https://openreview.net/forum?id=H1eA7AEtvS}
}

@misc{kaplan2020scalinglawsneurallanguage,
  title         = {Scaling Laws for Neural Language Models},
  author        = {Jared Kaplan and Sam McCandlish and Tom Henighan and Tom B. Brown and Benjamin Chess and Rewon Child and Scott Gray and Alec Radford and Jeffrey Wu and Dario Amodei},
  year          = {2020},
  eprint        = {2001.08361},
  archiveprefix = {arXiv},
  primaryclass  = {cs.LG},
  url           = {https://arxiv.org/abs/2001.08361}
}

@inproceedings{wang-etal-2024-scaling,
  title     = {Scaling Laws Across Model Architectures: A Comparative Analysis of Dense and {M}o{E} Models in Large Language Models},
  author    = {Wang, Siqi  and
               Chen, Zhengyu  and
               Li, Bei  and
               He, Keqing  and
               Zhang, Min  and
               Wang, Jingang},
  editor    = {Al-Onaizan, Yaser  and
               Bansal, Mohit  and
               Chen, Yun-Nung},
  booktitle = {Proceedings of the 2024 Conference on Empirical Methods in Natural Language Processing},
  month     = nov,
  year      = {2024},
  address   = {Miami, Florida, USA},
  publisher = {Association for Computational Linguistics},
  url       = {https://aclanthology.org/2024.emnlp-main.319/},
  doi       = {10.18653/v1/2024.emnlp-main.319},
  pages     = {5583--5595}
}

@misc{hoffmann2022trainingcomputeoptimallargelanguage,
  title         = {Training Compute-Optimal Large Language Models},
  author        = {Jordan Hoffmann and Sebastian Borgeaud and Arthur Mensch and Elena Buchatskaya and Trevor Cai and Eliza Rutherford and Diego de Las Casas and Lisa Anne Hendricks and Johannes Welbl and Aidan Clark and Tom Hennigan and Eric Noland and Katie Millican and George van den Driessche and Bogdan Damoc and Aurelia Guy and Simon Osindero and Karen Simonyan and Erich Elsen and Jack W. Rae and Oriol Vinyals and Laurent Sifre},
  year          = {2022},
  eprint        = {2203.15556},
  archiveprefix = {arXiv},
  primaryclass  = {cs.CL},
  url           = {https://arxiv.org/abs/2203.15556}
}

@inproceedings{NEURIPS2020_1457c0d6_gpt3,
  author    = {Brown, Tom and Mann, Benjamin and Ryder, Nick and Subbiah, Melanie and Kaplan, Jared D and Dhariwal, Prafulla and Neelakantan, Arvind and Shyam, Pranav and Sastry, Girish and Askell, Amanda and Agarwal, Sandhini and Herbert-Voss, Ariel and Krueger, Gretchen and Henighan, Tom and Child, Rewon and Ramesh, Aditya and Ziegler, Daniel and Wu, Jeffrey and Winter, Clemens and Hesse, Chris and Chen, Mark and Sigler, Eric and Litwin, Mateusz and Gray, Scott and Chess, Benjamin and Clark, Jack and Berner, Christopher and McCandlish, Sam and Radford, Alec and Sutskever, Ilya and Amodei, Dario},
  booktitle = {Advances in Neural Information Processing Systems},
  editor    = {H. Larochelle and M. Ranzato and R. Hadsell and M.F. Balcan and H. Lin},
  pages     = {1877--1901},
  publisher = {Curran Associates, Inc.},
  title     = {Language Models are Few-Shot Learners},
  url       = {https://proceedings.neurips.cc/paper_files/paper/2020/file/1457c0d6bfcb4967418bfb8ac142f64a-Paper.pdf},
  volume    = {33},
  year      = {2020}
}

@article{xie2025mhc,
  title   = {mHC: Manifold-Constrained Hyper-Connections},
  author  = {Xie, Zhenda and Wei, Yixuan and Cao, Huanqi and Zhao, Chenggang and Deng, Chengqi and Li, Jiashi and Dai, Damai and Gao, Huazuo and Chang, Jiang and Yu, Kuai and others},
  journal = {arXiv preprint arXiv:2512.24880},
  year    = {2025}
}

@article{sun2025curse,
  title   = {The Curse of Depth in Large Language Models},
  author  = {Sun, Wenfang and Song, Xinyuan and Li, Pengxiang and Yin, Lu and Zheng, Yefeng and Liu, Shiwei},
  journal = {arXiv preprint arXiv:2502.05795},
  year    = {2025}
}

@inproceedings{soldaini-etal-2024-dolma,
  title     = {Dolma: an Open Corpus of Three Trillion Tokens for Language Model Pretraining Research},
  author    = {Soldaini, Luca  and
               Kinney, Rodney  and
               Bhagia, Akshita  and
               Schwenk, Dustin  and
               Atkinson, David  and
               Authur, Russell  and
               Bogin, Ben  and
               Chandu, Khyathi  and
               Dumas, Jennifer  and
               Elazar, Yanai  and
               Hofmann, Valentin  and
               Jha, Ananya  and
               Kumar, Sachin  and
               Lucy, Li  and
               Lyu, Xinxi  and
               Lambert, Nathan  and
               Magnusson, Ian  and
               Morrison, Jacob  and
               Muennighoff, Niklas  and
               Naik, Aakanksha  and
               Nam, Crystal  and
               Peters, Matthew  and
               Ravichander, Abhilasha  and
               Richardson, Kyle  and
               Shen, Zejiang  and
               Strubell, Emma  and
               Subramani, Nishant  and
               Tafjord, Oyvind  and
               Walsh, Evan  and
               Zettlemoyer, Luke  and
               Smith, Noah  and
               Hajishirzi, Hannaneh  and
               Beltagy, Iz  and
               Groeneveld, Dirk  and
               Dodge, Jesse  and
               Lo, Kyle},
  editor    = {Ku, Lun-Wei  and
               Martins, Andre  and
               Srikumar, Vivek},
  booktitle = {Proceedings of the 62nd Annual Meeting of the Association for Computational Linguistics (Volume 1: Long Papers)},
  month     = aug,
  year      = {2024},
  address   = {Bangkok, Thailand},
  publisher = {Association for Computational Linguistics},
  url       = {https://aclanthology.org/2024.acl-long.840/},
  doi       = {10.18653/v1/2024.acl-long.840},
  pages     = {15725--15788}
}

@inproceedings{reid-etal-2022-m2d2,
  title     = {{M}2{D}2: A Massively Multi-Domain Language Modeling Dataset},
  author    = {Reid, Machel  and
               Zhong, Victor  and
               Gururangan, Suchin  and
               Zettlemoyer, Luke},
  editor    = {Goldberg, Yoav  and
               Kozareva, Zornitsa  and
               Zhang, Yue},
  booktitle = {Proceedings of the 2022 Conference on Empirical Methods in Natural Language Processing},
  month     = dec,
  year      = {2022},
  address   = {Abu Dhabi, United Arab Emirates},
  publisher = {Association for Computational Linguistics},
  url       = {https://aclanthology.org/2022.emnlp-main.63/},
  doi       = {10.18653/v1/2022.emnlp-main.63},
  pages     = {964--975}
}

@misc{merity2016pointer_wikitext,
  title         = {Pointer Sentinel Mixture Models},
  author        = {Stephen Merity and Caiming Xiong and James Bradbury and Richard Socher},
  year          = {2016},
  eprint        = {1609.07843},
  archiveprefix = {arXiv},
  primaryclass  = {cs.CL}
}

@article{allenai:arc,
  author  = {Peter Clark  and Isaac Cowhey and Oren Etzioni and Tushar Khot and
             Ashish Sabharwal and Carissa Schoenick and Oyvind Tafjord},
  title   = {Think you have Solved Question Answering? Try ARC, the AI2 Reasoning Challenge},
  journal = {arXiv:1803.05457v1},
  year    = {2018}
}

@inproceedings{zellers2019hellaswag,
  title     = {HellaSwag: Can a Machine Really Finish Your Sentence?},
  author    = {Zellers, Rowan and Holtzman, Ari and Bisk, Yonatan and Farhadi, Ali and Choi, Yejin},
  booktitle = {Proceedings of the 57th Annual Meeting of the Association for Computational Linguistics},
  year      = {2019}
}

@inproceedings{Bisk2020_piqa,
  author    = {Yonatan Bisk and Rowan Zellers and
               Ronan Le Bras and Jianfeng Gao
               and Yejin Choi},
  title     = {PIQA: Reasoning about Physical Commonsense in
               Natural Language},
  booktitle = {Thirty-Fourth AAAI Conference on
               Artificial Intelligence},
  year      = {2020}
}

@inproceedings{SciQ,
  title     = {Crowdsourcing Multiple Choice Science Questions},
  author    = {Welbl, Johannes  and
               Liu, Nelson F.  and
               Gardner, Matt},
  editor    = {Derczynski, Leon  and
               Xu, Wei  and
               Ritter, Alan  and
               Baldwin, Tim},
  booktitle = {Proceedings of the 3rd Workshop on Noisy User-generated Text},
  month     = sep,
  year      = {2017},
  address   = {Copenhagen, Denmark},
  publisher = {Association for Computational Linguistics},
  url       = {https://aclanthology.org/W17-4413/},
  doi       = {10.18653/v1/W17-4413},
  pages     = {94--106}
}

@inproceedings{talmor-etal-2019-commonsenseqa,
  title         = {{C}ommonsense{QA}: A Question Answering Challenge Targeting Commonsense Knowledge},
  author        = {Talmor, Alon  and
                   Herzig, Jonathan  and
                   Lourie, Nicholas  and
                   Berant, Jonathan},
  booktitle     = {Proceedings of the 2019 Conference of the North {A}merican Chapter of the Association for Computational Linguistics: Human Language Technologies, Volume 1 (Long and Short Papers)},
  month         = jun,
  year          = {2019},
  address       = {Minneapolis, Minnesota},
  publisher     = {Association for Computational Linguistics},
  url           = {https://aclanthology.org/N19-1421},
  doi           = {10.18653/v1/N19-1421},
  pages         = {4149--4158},
  archiveprefix = {arXiv},
  eprint        = {1811.00937},
  primaryclass  = {cs}
}

@article{2017arXivtriviaqa,
  author        = {{Joshi}, Mandar and {Choi}, Eunsol and {Weld},
                   Daniel and {Zettlemoyer}, Luke},
  title         = {{triviaqa: A Large Scale Distantly Supervised Challenge Dataset for Reading Comprehension}},
  journal       = {arXiv e-prints},
  year          = 2017,
  eid           = {arXiv:1705.03551},
  pages         = {arXiv:1705.03551},
  archiveprefix = {arXiv},
  eprint        = {1705.03551}
}

@article{kwiatkowski-etal-2019-naturalq,
  title     = {Natural Questions: A Benchmark for Question Answering Research},
  author    = {Kwiatkowski, Tom  and
               Palomaki, Jennimaria  and
               Redfield, Olivia  and
               Collins, Michael  and
               Parikh, Ankur  and
               Alberti, Chris  and
               Epstein, Danielle  and
               Polosukhin, Illia  and
               Devlin, Jacob  and
               Lee, Kenton  and
               Toutanova, Kristina  and
               Jones, Llion  and
               Kelcey, Matthew  and
               Chang, Ming-Wei  and
               Dai, Andrew M.  and
               Uszkoreit, Jakob  and
               Le, Quoc  and
               Petrov, Slav},
  editor    = {Lee, Lillian  and
               Johnson, Mark  and
               Roark, Brian  and
               Nenkova, Ani},
  journal   = {Transactions of the Association for Computational Linguistics},
  volume    = {7},
  year      = {2019},
  address   = {Cambridge, MA},
  publisher = {MIT Press},
  url       = {https://aclanthology.org/Q19-1026/},
  doi       = {10.1162/tacl_a_00276},
  pages     = {452--466}
}

@inproceedings{petty-etal-2024-impact,
  title     = {The Impact of Depth on Compositional Generalization in Transformer Language Models},
  author    = {Petty, Jackson  and
               Steenkiste, Sjoerd  and
               Dasgupta, Ishita  and
               Sha, Fei  and
               Garrette, Dan  and
               Linzen, Tal},
  editor    = {Duh, Kevin  and
               Gomez, Helena  and
               Bethard, Steven},
  booktitle = {Proceedings of the 2024 Conference of the North American Chapter of the Association for Computational Linguistics: Human Language Technologies (Volume 1: Long Papers)},
  month     = jun,
  year      = {2024},
  address   = {Mexico City, Mexico},
  publisher = {Association for Computational Linguistics},
  url       = {https://aclanthology.org/2024.naacl-long.402/},
  doi       = {10.18653/v1/2024.naacl-long.402},
  pages     = {7239--7252}
}

@misc{grattafiori2024llama3herdmodels,
  title         = {The Llama 3 Herd of Models},
  author        = {Aaron Grattafiori and Abhimanyu Dubey and Abhinav Jauhri and Abhinav Pandey and Abhishek Kadian and Ahmad Al-Dahle and Aiesha Letman and Akhil Mathur and Alan Schelten and Alex Vaughan and Amy Yang and Angela Fan and Anirudh Goyal and Anthony Hartshorn and Aobo Yang and Archi Mitra and Archie Sravankumar and Artem Korenev and Arthur Hinsvark and Arun Rao and Aston Zhang and Aurelien Rodriguez and Austen Gregerson and Ava Spataru and Baptiste Roziere and Bethany Biron and Binh Tang and Bobbie Chern and Charlotte Caucheteux and Chaya Nayak and Chloe Bi and Chris Marra and Chris McConnell and Christian Keller and Christophe Touret and Chunyang Wu and Corinne Wong and Cristian Canton Ferrer and Cyrus Nikolaidis and Damien Allonsius and Daniel Song and Danielle Pintz and Danny Livshits and Danny Wyatt and David Esiobu and Dhruv Choudhary and Dhruv Mahajan and Diego Garcia-Olano and Diego Perino and Dieuwke Hupkes and Egor Lakomkin and Ehab AlBadawy and Elina Lobanova and Emily Dinan and Eric Michael Smith and Filip Radenovic and Francisco Guzmán and Frank Zhang and Gabriel Synnaeve and Gabrielle Lee and Georgia Lewis Anderson and Govind Thattai and Graeme Nail and Gregoire Mialon and Guan Pang and Guillem Cucurell and Hailey Nguyen and Hannah Korevaar and Hu Xu and Hugo Touvron and Iliyan Zarov and Imanol Arrieta Ibarra and Isabel Kloumann and Ishan Misra and Ivan Evtimov and Jack Zhang and Jade Copet and Jaewon Lee and Jan Geffert and Jana Vranes and Jason Park and Jay Mahadeokar and Jeet Shah and Jelmer van der Linde and Jennifer Billock and Jenny Hong and Jenya Lee and Jeremy Fu and Jianfeng Chi and Jianyu Huang and Jiawen Liu and Jie Wang and Jiecao Yu and Joanna Bitton and Joe Spisak and Jongsoo Park and Joseph Rocca and Joshua Johnstun and Joshua Saxe and Junteng Jia and Kalyan Vasuden Alwala and Karthik Prasad and Kartikeya Upasani and Kate Plawiak and Ke Li and Kenneth Heafield and Kevin Stone and Khalid El-Arini and Krithika Iyer and Kshitiz Malik and Kuenley Chiu and Kunal Bhalla and Kushal Lakhotia and Lauren Rantala-Yeary and Laurens van der Maaten and Lawrence Chen and Liang Tan and Liz Jenkins and Louis Martin and Lovish Madaan and Lubo Malo and Lukas Blecher and Lukas Landzaat and Luke de Oliveira and Madeline Muzzi and Mahesh Pasupuleti and Mannat Singh and Manohar Paluri and Marcin Kardas and Maria Tsimpoukelli and Mathew Oldham and Mathieu Rita and Maya Pavlova and Melanie Kambadur and Mike Lewis and Min Si and Mitesh Kumar Singh and Mona Hassan and Naman Goyal and Narjes Torabi and Nikolay Bashlykov and Nikolay Bogoychev and Niladri Chatterji and Ning Zhang and Olivier Duchenne and Onur Çelebi and Patrick Alrassy and Pengchuan Zhang and Pengwei Li and Petar Vasic and Peter Weng and Prajjwal Bhargava and Pratik Dubal and Praveen Krishnan and Punit Singh Koura and Puxin Xu and Qing He and Qingxiao Dong and Ragavan Srinivasan and Raj Ganapathy and Ramon Calderer and Ricardo Silveira Cabral and Robert Stojnic and Roberta Raileanu and Rohan Maheswari and Rohit Girdhar and Rohit Patel and Romain Sauvestre and Ronnie Polidoro and Roshan Sumbaly and Ross Taylor and Ruan Silva and Rui Hou and Rui Wang and Saghar Hosseini and Sahana Chennabasappa and Sanjay Singh and Sean Bell and Seohyun Sonia Kim and Sergey Edunov and Shaoliang Nie and Sharan Narang and Sharath Raparthy and Sheng Shen and Shengye Wan and Shruti Bhosale and Shun Zhang and Simon Vandenhende and Soumya Batra and Spencer Whitman and Sten Sootla and Stephane Collot and Suchin Gururangan and Sydney Borodinsky and Tamar Herman and Tara Fowler and Tarek Sheasha and Thomas Georgiou and Thomas Scialom and Tobias Speckbacher and Todor Mihaylov and Tong Xiao and Ujjwal Karn and Vedanuj Goswami and Vibhor Gupta and Vignesh Ramanathan and Viktor Kerkez and Vincent Gonguet and Virginie Do and Vish Vogeti and Vítor Albiero and Vladan Petrovic and Weiwei Chu and Wenhan Xiong and Wenyin Fu and Whitney Meers and Xavier Martinet and Xiaodong Wang and Xiaofang Wang and Xiaoqing Ellen Tan and Xide Xia and Xinfeng Xie and Xuchao Jia and Xuewei Wang and Yaelle Goldschlag and Yashesh Gaur and Yasmine Babaei and Yi Wen and Yiwen Song and Yuchen Zhang and Yue Li and Yuning Mao and Zacharie Delpierre Coudert and Zheng Yan and Zhengxing Chen and Zoe Papakipos and Aaditya Singh and Aayushi Srivastava and Abha Jain and Adam Kelsey and Adam Shajnfeld and Adithya Gangidi and Adolfo Victoria and Ahuva Goldstand and Ajay Menon and Ajay Sharma and Alex Boesenberg and Alexei Baevski and Allie Feinstein and Amanda Kallet and Amit Sangani and Amos Teo and Anam Yunus and Andrei Lupu and Andres Alvarado and Andrew Caples and Andrew Gu and Andrew Ho and Andrew Poulton and Andrew Ryan and Ankit Ramchandani and Annie Dong and Annie Franco and Anuj Goyal and Aparajita Saraf and Arkabandhu Chowdhury and Ashley Gabriel and Ashwin Bharambe and Assaf Eisenman and Azadeh Yazdan and Beau James and Ben Maurer and Benjamin Leonhardi and Bernie Huang and Beth Loyd and Beto De Paola and Bhargavi Paranjape and Bing Liu and Bo Wu and Boyu Ni and Braden Hancock and Bram Wasti and Brandon Spence and Brani Stojkovic and Brian Gamido and Britt Montalvo and Carl Parker and Carly Burton and Catalina Mejia and Ce Liu and Changhan Wang and Changkyu Kim and Chao Zhou and Chester Hu and Ching-Hsiang Chu and Chris Cai and Chris Tindal and Christoph Feichtenhofer and Cynthia Gao and Damon Civin and Dana Beaty and Daniel Kreymer and Daniel Li and David Adkins and David Xu and Davide Testuggine and Delia David and Devi Parikh and Diana Liskovich and Didem Foss and Dingkang Wang and Duc Le and Dustin Holland and Edward Dowling and Eissa Jamil and Elaine Montgomery and Eleonora Presani and Emily Hahn and Emily Wood and Eric-Tuan Le and Erik Brinkman and Esteban Arcaute and Evan Dunbar and Evan Smothers and Fei Sun and Felix Kreuk and Feng Tian and Filippos Kokkinos and Firat Ozgenel and Francesco Caggioni and Frank Kanayet and Frank Seide and Gabriela Medina Florez and Gabriella Schwarz and Gada Badeer and Georgia Swee and Gil Halpern and Grant Herman and Grigory Sizov and Guangyi and Zhang and Guna Lakshminarayanan and Hakan Inan and Hamid Shojanazeri and Han Zou and Hannah Wang and Hanwen Zha and Haroun Habeeb and Harrison Rudolph and Helen Suk and Henry Aspegren and Hunter Goldman and Hongyuan Zhan and Ibrahim Damlaj and Igor Molybog and Igor Tufanov and Ilias Leontiadis and Irina-Elena Veliche and Itai Gat and Jake Weissman and James Geboski and James Kohli and Janice Lam and Japhet Asher and Jean-Baptiste Gaya and Jeff Marcus and Jeff Tang and Jennifer Chan and Jenny Zhen and Jeremy Reizenstein and Jeremy Teboul and Jessica Zhong and Jian Jin and Jingyi Yang and Joe Cummings and Jon Carvill and Jon Shepard and Jonathan McPhie and Jonathan Torres and Josh Ginsburg and Junjie Wang and Kai Wu and Kam Hou U and Karan Saxena and Kartikay Khandelwal and Katayoun Zand and Kathy Matosich and Kaushik Veeraraghavan and Kelly Michelena and Keqian Li and Kiran Jagadeesh and Kun Huang and Kunal Chawla and Kyle Huang and Lailin Chen and Lakshya Garg and Lavender A and Leandro Silva and Lee Bell and Lei Zhang and Liangpeng Guo and Licheng Yu and Liron Moshkovich and Luca Wehrstedt and Madian Khabsa and Manav Avalani and Manish Bhatt and Martynas Mankus and Matan Hasson and Matthew Lennie and Matthias Reso and Maxim Groshev and Maxim Naumov and Maya Lathi and Meghan Keneally and Miao Liu and Michael L. Seltzer and Michal Valko and Michelle Restrepo and Mihir Patel and Mik Vyatskov and Mikayel Samvelyan and Mike Clark and Mike Macey and Mike Wang and Miquel Jubert Hermoso and Mo Metanat and Mohammad Rastegari and Munish Bansal and Nandhini Santhanam and Natascha Parks and Natasha White and Navyata Bawa and Nayan Singhal and Nick Egebo and Nicolas Usunier and Nikhil Mehta and Nikolay Pavlovich Laptev and Ning Dong and Norman Cheng and Oleg Chernoguz and Olivia Hart and Omkar Salpekar and Ozlem Kalinli and Parkin Kent and Parth Parekh and Paul Saab and Pavan Balaji and Pedro Rittner and Philip Bontrager and Pierre Roux and Piotr Dollar and Polina Zvyagina and Prashant Ratanchandani and Pritish Yuvraj and Qian Liang and Rachad Alao and Rachel Rodriguez and Rafi Ayub and Raghotham Murthy and Raghu Nayani and Rahul Mitra and Rangaprabhu Parthasarathy and Raymond Li and Rebekkah Hogan and Robin Battey and Rocky Wang and Russ Howes and Ruty Rinott and Sachin Mehta and Sachin Siby and Sai Jayesh Bondu and Samyak Datta and Sara Chugh and Sara Hunt and Sargun Dhillon and Sasha Sidorov and Satadru Pan and Saurabh Mahajan and Saurabh Verma and Seiji Yamamoto and Sharadh Ramaswamy and Shaun Lindsay and Shaun Lindsay and Sheng Feng and Shenghao Lin and Shengxin Cindy Zha and Shishir Patil and Shiva Shankar and Shuqiang Zhang and Shuqiang Zhang and Sinong Wang and Sneha Agarwal and Soji Sajuyigbe and Soumith Chintala and Stephanie Max and Stephen Chen and Steve Kehoe and Steve Satterfield and Sudarshan Govindaprasad and Sumit Gupta and Summer Deng and Sungmin Cho and Sunny Virk and Suraj Subramanian and Sy Choudhury and Sydney Goldman and Tal Remez and Tamar Glaser and Tamara Best and Thilo Koehler and Thomas Robinson and Tianhe Li and Tianjun Zhang and Tim Matthews and Timothy Chou and Tzook Shaked and Varun Vontimitta and Victoria Ajayi and Victoria Montanez and Vijai Mohan and Vinay Satish Kumar and Vishal Mangla and Vlad Ionescu and Vlad Poenaru and Vlad Tiberiu Mihailescu and Vladimir Ivanov and Wei Li and Wenchen Wang and Wenwen Jiang and Wes Bouaziz and Will Constable and Xiaocheng Tang and Xiaojian Wu and Xiaolan Wang and Xilun Wu and Xinbo Gao and Yaniv Kleinman and Yanjun Chen and Ye Hu and Ye Jia and Ye Qi and Yenda Li and Yilin Zhang and Ying Zhang and Yossi Adi and Youngjin Nam and Yu and Wang and Yu Zhao and Yuchen Hao and Yundi Qian and Yunlu Li and Yuzi He and Zach Rait and Zachary DeVito and Zef Rosnbrick and Zhaoduo Wen and Zhenyu Yang and Zhiwei Zhao and Zhiyu Ma},
  year          = {2024},
  eprint        = {2407.21783},
  archiveprefix = {arXiv},
  primaryclass  = {cs.AI},
  url           = {https://arxiv.org/abs/2407.21783}
}

@inproceedings{mcleish2025gemstones,
  title     = {Gemstones: A Model Suite for Multi-Faceted Scaling Laws},
  author    = {Sean Michael McLeish and John Kirchenbauer and David Yu Miller and Siddharth Singh and Abhinav Bhatele and Micah Goldblum and Ashwinee Panda and Tom Goldstein},
  booktitle = {The Thirty-ninth Annual Conference on Neural Information Processing Systems},
  year      = {2025},
  url       = {https://openreview.net/forum?id=iZk78dZ1Ap}
}

@misc{godey2026lostbackpropagationlmhead,
  title         = {Lost in Backpropagation: The LM Head is a Gradient Bottleneck},
  author        = {Nathan Godey and Yoav Artzi},
  year          = {2026},
  eprint        = {2603.10145},
  archiveprefix = {arXiv},
  primaryclass  = {cs.CL},
  url           = {https://arxiv.org/abs/2603.10145}
}

@article{Johnson1984ExtensionsOL,
  title   = {Extensions of Lipschitz mappings into Hilbert space},
  author  = {William B. Johnson and Joram Lindenstrauss},
  journal = {Contemporary mathematics},
  year    = {1984},
  volume  = {26},
  pages   = {189-206},
  url     = {https://api.semanticscholar.org/CorpusID:117819162}
}

@article{dasgupta2003elementary,
  author   = {Dasgupta, Sanjoy and Gupta, Anupam},
  title    = {An elementary proof of a theorem of Johnson and Lindenstrauss},
  journal  = {Random Structures \& Algorithms},
  volume   = {22},
  number   = {1},
  pages    = {60-65},
  doi      = {https://doi.org/10.1002/rsa.10073},
  url      = {https://onlinelibrary.wiley.com/doi/abs/10.1002/rsa.10073},
  eprint   = {https://onlinelibrary.wiley.com/doi/pdf/10.1002/rsa.10073},
  year     = {2003}
}

@inproceedings{
Rae2020Compressive,
title={Compressive Transformers for Long-Range Sequence Modelling},
author={Jack W. Rae and Anna Potapenko and Siddhant M. Jayakumar and Chloe Hillier and Timothy P. Lillicrap},
booktitle={International Conference on Learning Representations},
year={2020},
url={https://openreview.net/forum?id=SylKikSYDH}
}

@misc{proofpile2022,
  title         = {Proof-pile},
  author        = {Zhangir Azerbayev and Edward Ayers and Bartosz Piotrowski},
  year          = {2022},
  url           = {https://github.com/zhangir-azerbayev/proof-pile}
}

@inproceedings{
hsieh2024ruler,
title={{RULER}: What{\textquoteright}s the Real Context Size of Your Long-Context Language Models?},
author={Cheng-Ping Hsieh and Simeng Sun and Samuel Kriman and Shantanu Acharya and Dima Rekesh and Fei Jia and Boris Ginsburg},
booktitle={First Conference on Language Modeling},
year={2024},
url={https://openreview.net/forum?id=kIoBbc76Sy}
}

\newpage
\appendix

% You can have as much text here as you want. The main body must be at most $8$
% pages long. For the final version, one more page can be added. If you want, you
% can use an appendix like this one.

% The $\mathtt{\backslash onecolumn}$ command above can be kept in place if you
% prefer a one-column appendix, or can be removed if you prefer a two-column
% appendix.  Apart from this possible change, the style (font size, spacing,
% margins, page numbering, etc.) should be kept the same as the main body.

\section{Appendix}
\subsection{Experimental Details}
\label{sec:experiment_setting_details}

\subsubsection{Experimental Environment and Training Corpus}
Training is conducted with NVIDIA NeMo Megatron Bridge\footnote{\href{https://github.com/NVIDIA-NeMo/Megatron-Bridge}{https://github.com/NVIDIA-NeMo/Megatron-Bridge}}, while evaluation uses the OLMo codebase\footnote{\href{https://github.com/allenai/OLMo}{https://github.com/allenai/OLMo}}. Experiments are executed on NVIDIA RTX 6000 Ada and B200 GPUs. The 3B and 8B runs use 8 B200 GPUs and take approximately 3 and 8 days, respectively. The fixed random seed 6198 is used in all experiments.

For training data, we adopt the Stage 1 pre-training corpus used to train the original OLMo-2 1B checkpoint. To control for the effects of stochasticity arising from data shuffling, we replicate the exact data ordering employed during the Stage 1 pre-training of OLMo-2 1B. From this fixed-order dataset, we select the first 2.5B and 7B tokens for the 113M and 403M studies, respectively. The token budgets for the 906M, 3B, and 8B scaling experiments are reported in Table~\ref{tab:exp_sum}. The choice of the number of training tokens for each model size follows \citep{hoffmann2022trainingcomputeoptimallargelanguage} reported ratio, giving approximately 20 tokens per parameter at every scale.

The finetuning and evaluation codebase\footnote{\href{https://github.com/jquesnelle/yarn}{https://github.com/jquesnelle/yarn}} of the long-context experiments follows \citep{peng2023yarnefficientcontextwindow} with necessary adaptation to use the NeMo checkpoints and model. We convert the NeMo checkpoints to HuggingFace format for finetuning and evaluation. For finetuning and evaluation, we employ the exact same hyperparameters as \citep{peng2023yarnefficientcontextwindow}. The finetuning corpus follows \citep{peng2023yarnefficientcontextwindow}, which employs PG19 from \citep{Rae2020Compressive}, and the evaluation is performed on Proof-pile from \citep{proofpile2022}. Both finetuning and evaluation dataset are processed with the same tokenizer as the training data, which is the tokenizer used in OLMo-2 1B model.
% For validation, we use the \texttt{c4\_en-validation} dataset provided in the OLMo codebase as the primary validation set. All validation loss and perplexity values reported in the tables are computed on this dataset.

\subsubsection{Validation Datasets}

We conduct validation using the following datasets: Dolma Common Crawl \cite{soldaini-etal-2024-dolma}, Dolma The Stack \cite{soldaini-etal-2024-dolma}, M2D2 \cite{reid-etal-2022-m2d2}, and WikiText \cite{merity2016pointer_wikitext}. The validation splits follow the official OLMo-2 1B configuration. For both cross-entropy loss and perplexity, we report the average scores computed across these four datasets. Additional validation results are provided in Section~\ref{sec:validation_results}.

\subsubsection{Downstream Evaluation Datasets}

We conduct downstream task evaluation on the following benchmark datasets: Arc Easy \cite{allenai:arc}, HellaSwag \cite{zellers2019hellaswag}, PIQA \cite{Bisk2020_piqa}, SciQ \cite{SciQ}, CommonsenseQA \cite{talmor-etal-2019-commonsenseqa}, TriviaQA \cite{2017arXivtriviaqa}, and NaturalQS \cite{kwiatkowski-etal-2019-naturalq}. Among these, TriviaQA and NaturalQS are evaluated using perplexity, while all remaining tasks are assessed using accuracy as the evaluation metric.

\subsubsection{Training Hyperparameters}

Across all experiments, we use AdamW as the optimizer and adopt the cross-entropy loss augmented with an auxiliary softmax loss as the training objective. For AdamW, the hyperparameters are set to $\beta_1 = 0.9$, $\beta_2 = 0.95$, and $\epsilon = 1 \times 10^{-8}$, with a weight decay coefficient of $\lambda = 0.1$. SwiGLU is employed as the activation function for all models. Rotary Position Embedding (RoPE) is utilized for positional encoding in both the Transformer architecture and the proposed hourglass blocks. A cosine learning rate scheduler with warmup is applied in all experiments, where the number of warmup tokens varies according to model size. Additional hyperparameters are reported in Table~\ref{tab:exp_sum}, and model-size-dependent architectural configurations are reported in Table~\ref{tab:baseline_and_hourglass_configs}.

% \begin{table}[h!]
% \vspace{+3mm}
%     \centering
%     \caption{\textbf{Additional experimental hyperparameters for different model sizes}, including learning rates, warmup tokens, attention heads, and sequence lengths.}
%     \label{tab:exp_sum}
%     \resizebox{0.7\linewidth}{!}{
%     \begin{tabular}{l|ccccc}
%         \toprule
%         Model Size & Learning Rate & Warm\-Up Tokens & Attention Heads & Max Sequence Size \\
%         \midrule
%         113M & $6\times 10^{-4}$ & 50M & 12 & 2048 \\
%         403M & $3\times 10^{-4}$ & 50M & 16 & 2048 \\
%         906M & $2.5\times 10^{-4}$ & 50M & 16 & 2048 \\
%         1074M & $4\times 10^{-4}$ & 200M & 16 & 4096 \\
%         \bottomrule
%     \end{tabular}
%     }
%     \vspace{+3mm}
% \end{table}
\begin{table}[ht!]
    \vspace{+3mm}
    \centering
    \caption{\textbf{Additional experimental hyperparameters for different model sizes}, including learning rates, batch tokens, trained tokens, warmup tokens, attention heads, and sequence lengths.}
    \label{tab:exp_sum}
    \resizebox{0.9\linewidth}{!}{
        \begin{tabular}{l|cccccc}
            \toprule
            Model Size & Learning Rate       & Batch Tokens (K) & Trained Tokens (B) & Warm\-Up Tokens & Attention Heads & Max Sequence Size \\
            \midrule
            113M       & $6\times 10^{-4}$   & 512              & 2.5                & 50M             & 12              & 2048              \\
            403M       & $3\times 10^{-4}$   & 1024             & 7                  & 50M             & 16              & 4096              \\
            906M       & $2.5\times 10^{-4}$ & 1024             & 18                 & 100M            & 16              & 4096              \\
            3B         & $1.6\times 10^{-4}$ & 1024             & 60                 & 400M            & 24              & 4096              \\
            8B         & $1.2\times 10^{-4}$ & 1024             & 160                & 400M            & 32              & 4096              \\
            \bottomrule
        \end{tabular}
    }
    \vspace{+3mm}
\end{table}

\subsection{Conventional Baseline Controls}
\label{sec:appendix_model_efficiency}

We use standard published Transformer configurations as the primary baselines because they provide fixed, reproducible reference points for the established conventional design.
As targeted controls, \autoref{tab:baseline_hpsearch_ratio} evaluates two classes of conventional alternatives.
At 403M, varying the conventional FFN ratio yields similar quality, while the standard $d_h/d_{\text{model}}=4$ ratio has the lowest training FLOPs among these comparable variants.
At 906M, a reduced-depth conventional variant produces a mixed trade-off: the parameter-matched $L=12$ variant reduces training FLOPs by about $10\%$, but it has worse validation PPL and average downstream PPL.
These checks support the use of standard configurations as fair primary baselines, while a full conventional depth--width frontier remains outside the scope of this study.

\begin{table*}[t!]
    \centering
    \caption{\textbf{Conventional baseline controls.} Results for conventional baseline models with different internal FFN ratios and an exploratory reduced-depth 906M variant. Arrows indicate the preferred direction for each metric. Parenthesized values report the relative change versus the standard setting at the same parameter scale. The reduced-depth variant illustrates that reallocating the same parameter budget into fewer conventional layers can lower training FLOPs while worsening validation PPL and average downstream PPL.}
    \label{tab:baseline_hpsearch_ratio}
    \setlength{\tabcolsep}{2pt}
    \renewcommand{\arraystretch}{0.85}
    \scriptsize
    \newcommand{\baselinedelta}[1]{{\tiny #1}}
    \resizebox{0.98\linewidth}{!}{
        \begin{tabular}{lllll|ll|ll}
            \toprule
            Params & Setting   & L  & $d_h/d_{\text{model}}$ & FLOPs $\downarrow$                    & Val Loss $\downarrow$             & Val PPL $\downarrow$              & Avg Downstream Acc $\uparrow$     & Avg Downstream PPL $\downarrow$   \\
            \midrule
            403M   & Standard  & 24 & 4                      & 1.87e+15                              & 3.081                             & 25.21                             & 0.525                             & 1.472                             \\
            403M   & FFN-ratio & 24 & 3                      & 1.93e+15 \baselinedelta{($+3.21\%$)}  & 3.081 \baselinedelta{($+0.00\%$)} & 25.30 \baselinedelta{($+0.36\%$)} & 0.525 \baselinedelta{($+0.00\%$)} & 1.503 \baselinedelta{($+2.11\%$)} \\
            403M   & FFN-ratio & 24 & 2                      & 2.04e+15 \baselinedelta{($+9.09\%$)}  & 3.067 \baselinedelta{($-0.45\%$)} & 24.97 \baselinedelta{($-0.95\%$)} & 0.523 \baselinedelta{($-0.38\%$)} & 1.486 \baselinedelta{($+0.95\%$)} \\
            403M   & FFN-ratio & 24 & 1.5                    & 2.10e+15 \baselinedelta{($+12.30\%$)} & 3.060 \baselinedelta{($-0.68\%$)} & 24.80 \baselinedelta{($-1.63\%$)} & 0.528 \baselinedelta{($+0.57\%$)} & 1.518 \baselinedelta{($+3.12\%$)} \\
            \midrule
            906M   & Standard  & 24 & 4                      & 3.34e+16                              & 3.092                             & 26.00                             & 0.503                             & 1.524                             \\
            906M   & FFN-ratio & 12 & 9.33                   & 3.01e+16 \baselinedelta{($-9.88\%$)}  & 3.127 \baselinedelta{($+1.13\%$)} & 26.85 \baselinedelta{($+3.27\%$)} & 0.503 \baselinedelta{($+0.00\%$)} & 1.578 \baselinedelta{($+3.54\%$)} \\
            \bottomrule
        \end{tabular}
    }
\end{table*}

% Detailed downstream metrics for Table~\ref{tab:baseline_hpsearch_ratio}.
% \begin{table*}[t!]
%     \centering
%     \caption{\textbf{Detailed downstream metrics for conventional baseline controls.}}
%     \label{tab:baseline_hpsearch_ratio_detailed_commented}
%     \setlength{\tabcolsep}{2pt}
%     \renewcommand{\arraystretch}{0.85}
%     \scriptsize
%     \resizebox{1\linewidth}{!}{
%         \begin{tabular}{clcll|ccccc|cc}
%             \toprule
%             Params & Setting & L  & $d_h/d_{\text{model}}$ & TFLOPs & Arc Easy & HellaSwag & PIQA & SciQ & CommonsenseQA & TriviaQA (PPL) & NaturalQS (PPL) \\
%             \midrule
%             403M & Standard     & 24 & 4    & 1.87e+15 & 0.505 & 0.356 & 0.656 & 0.770 & 0.337 & 1.572 & 1.372 \\
%             403M & FFN-ratio    & 24 & 3    & 1.93e+15 & 0.516 & 0.352 & 0.657 & 0.764 & 0.335 & 1.598 & 1.409 \\
%             403M & FFN-ratio    & 24 & 2    & 2.04e+15 & 0.495 & 0.352 & 0.667 & 0.772 & 0.328 & 1.599 & 1.373 \\
%             403M & FFN-ratio    & 24 & 1.5  & 2.10e+15 & 0.521 & 0.354 & 0.649 & 0.790 & 0.328 & 1.606 & 1.430 \\
%             \midrule
%             906M & Standard     & 24 & 4    & 3.34e+16 & 0.504 & 0.331 & 0.643 & 0.735 & 0.301 & 1.645 & 1.404 \\
%             906M & Param-match  & 12 & 9.33 & 3.01e+16 & 0.500 & 0.330 & 0.646 & 0.731 & 0.307 & 1.712 & 1.444 \\
%             \bottomrule
%         \end{tabular}
%     }
% \end{table*}

\subsection{Detailed Validation Results}
\label{sec:validation_results}

\begin{table}[ht!]
    \centering
    \caption{\textbf{Detailed validation loss and perplexity on our 4 validation datasets}, including Dolma Common Crawl, Dolma The Stack, M2D2 and WikiText.}
    \label{tab:detailed_val}

    \resizebox{1\linewidth}{!}{
        \begin{tabular}{l|l|cccc|ccccc}
            \toprule
                                  &                                           & \multicolumn{4}{c|}{Validation Loss}
                                  & \multicolumn{4}{c}{Validation Perplexity}                                                           \\
            \cmidrule(lr){3-6} \cmidrule(lr){7-10}
            Size                  & Model
                                  & Dolma Common Crawl                        & Dolma The Stack                      & M2D2  & WikiText
                                  & Dolma Common Crawl                        & Dolma The Stack                      & M2D2  & WikiText \\
            \midrule

            \multirow{2}{*}{906M} & Conventional
                                  & 3.554                                     & 1.981                                & 3.581 & 3.253
                                  & 34.95                                     & 7.25                                 & 35.93 & 25.86    \\

                                  & Hourglass
                                  & 3.555                                     & 1.984                                & 3.582 & 3.260
                                  & 34.98                                     & 7.27                                 & 35.96 & 26.04
            \\
            \midrule

            \multirow{2}{*}{3B}   & Conventional
                                  & 3.139                                     & 1.545                                & 3.201 & 2.696
                                  & 23.07                                     & 4.69                                 & 24.60 & 14.82    \\

                                  & Hourglass
                                  & 3.140                                     & 1.546                                & 3.225 & 2.694
                                  & 23.10                                     & 4.69                                 & 25.15 & 14.80
            \\
            % \midrule
            %
            % \multirow{2}{*}{8B}   & Conventional
            %                       & 2.980                                     & 1.405                                & 3.079 & 2.493
            %                       & 19.68                                     & 4.07                                 & 21.74 & 12.10    \\
            %
            %                       & Hourglass
            %                       & \TBD                                      & \TBD                                 & \TBD  & \TBD
            %                       & \TBD                                      & \TBD                                 & \TBD  & \TBD
            % \\
            \bottomrule
        \end{tabular}
    }
\end{table}

Table~\ref{tab:detailed_val} presents the complete validation loss and perplexity results for the 906M and 3B scaling experiments evaluated across four distinct validation datasets.

\begin{table*}[t!]
    \centering
    \caption{\textbf{Full per-task performance for the 906M, 3B, and 8B scaling experiments.} The main text reports the relative deltas; this table provides the raw validation and downstream metrics.}
    \label{tab:appendix_scale_results}
    \setlength{\tabcolsep}{3pt}
    \renewcommand{\arraystretch}{0.85}
    \footnotesize
    \resizebox{1\linewidth}{!}{
        \begin{tabular}{l|l|cc|ccccc|cc|cc}
            \toprule
                                  &                                     & \multicolumn{2}{c|}{Validation}
                                  & \multicolumn{5}{c|}{Accuracy Tasks} & \multicolumn{2}{c|}{PPL Tasks}  & \multicolumn{2}{c}{Avg}                                   \\
            \cmidrule(lr){3-4} \cmidrule(lr){5-9} \cmidrule(lr){10-11} \cmidrule(lr){12-13}
            Size                  & Model
                                  & Val Loss                            & Val PPL
                                  & Arc Easy                            & HellaSwag                       & PIQA                    & SciQ
                                  & CommonsenseQA                       & TriviaQA                        & NaturalQS               & Avg Acc        & Avg PPL        \\
            \midrule

            \multirow{2}{*}{906M} & Conventional
                                  & \textbf{3.092}                      & \textbf{26.00}
                                  & \textbf{0.504}                      & 0.331                           & 0.643                   & \textbf{0.735}
                                  & 0.301                               & \textbf{1.645}                  & 1.404                   & 0.503          & 1.524          \\

                                  & Hourglass
                                  & 3.095                               & 26.06
                                  & 0.498                               & \textbf{0.337}                  & \textbf{0.654}          & 0.727
                                  & \textbf{0.309}                      & 1.657                           & \textbf{1.372}          & \textbf{0.505} & \textbf{1.514} \\
            \midrule

            \multirow{2}{*}{3B}   & Conventional
                                  & \textbf{2.646}                      & \textbf{16.80}
                                  & 0.628                               & \textbf{0.519}                  & \textbf{0.726}          & \textbf{0.859}
                                  & 0.399                               & 1.233                           & 1.190                   & 0.626          & 1.211          \\

                                  & Hourglass
                                  & 2.651                               & 16.93
                                  & \textbf{0.654}                      & 0.517                           & 0.716                   & 0.852
                                  & \textbf{0.402}                      & \textbf{1.215}                  & \textbf{1.161}          & \textbf{0.628} & \textbf{1.188} \\
            \midrule

            \multirow{2}{*}{8B}   & Conventional
                                  & 2.414                               & 13.37
                                  & \textbf{0.723}                      & \textbf{0.667}                  & 0.754                   & 0.911
                                  & 0.453                               & 0.909                           & 1.000                   & 0.701          & 0.955          \\

                                  & Hourglass
                                  & \textbf{2.409}                      & \textbf{13.31}
                                  & 0.721                               & 0.661                           & \textbf{0.764}          & \textbf{0.921}
                                  & \textbf{0.462}                      & \textbf{0.900}                  & \textbf{0.986}          & \textbf{0.706} & \textbf{0.943} \\

            \bottomrule
        \end{tabular}
    }
    \vspace{-2mm}
\end{table*}

\subsection{Sensitivity to Reduced Attention Dimension}
\label{sec:attn_reduction}

Random projection analyses provide useful context: geometric structure can be preserved in lower-dimensional projections when the target dimension is sufficiently large~\citep{Johnson1984ExtensionsOL,dasgupta2003elementary}.
The Hourglass architecture operates at a wider $d_{\text{model}}$ than the conventional baseline, so attention queries and keys are projected from a higher-dimensional residual stream into the same attention width as the baseline.
As a small sensitivity check, we reduce the attention dimension $d_{\text{attn}}$ to $0.9\times$ the default value at 906M parameters and compare the resulting degradation in \autoref{tab:attn_reduction}.
Reducing $d_{\text{attn}}$ increases validation PPL by $+0.43$ for the conventional baseline and by $+0.21$ for the Hourglass model, with a similar pattern in validation loss.
This shows lower degradation for the Hourglass variant under this moderate attention-width compression check, although this experiment is not intended as a complete attention-width sweep.

\begin{table*}[t!]
    \centering
    \caption{\textbf{Sensitivity to reduced $d_{\text{attn}}$ ($0.9\times$) at 906M parameters.} The $\Delta$ columns show the degradation from reducing $d_{\text{attn}}$. In this setting, the Hourglass model shows roughly half the validation-loss and perplexity degradation of the conventional baseline.}
    \label{tab:attn_reduction}
    \setlength{\tabcolsep}{5pt}
    \renewcommand{\arraystretch}{0.95}
    \footnotesize
    \resizebox{0.85\linewidth}{!}{
        \begin{tabular}{l|cc|cc|cc|cc}
            \toprule
            Model              & $d_{\text{model}}$ & $d_{\text{attn}}$ & Params & FLOPs    & Val Loss & $\Delta$                & Val PPL & $\Delta$               \\
            \midrule
            Conventional       & 1536               & 1536              & 906.1M & 3.34e+16 & 3.092    &                         & 26.00   &                        \\
            Conventional (0.9) & 1536               & 1376              & 882.5M & 3.21e+16 & 3.110    & \textcolor{red}{+0.018} & 26.43   & \textcolor{red}{+0.43} \\
            \midrule
            Hourglass          & 2496               & 1536              & 908.8M & 3.26e+16 & 3.095    &                         & 26.06   &                        \\
            Hourglass (0.9)    & 2496               & 1376              & 889.6M & 3.17e+16 & 3.103    & +0.008                  & 26.27   & +0.21                  \\
            \bottomrule
        \end{tabular}
    }
\end{table*}

\subsection{No-Attention FFN Control}
\label{sec:appendix_bigram}

As a minimal FFN-only control, we train 113M-parameter models with all self-attention layers removed, reducing the architecture to a token-level feed-forward network.
This experiment is intentionally limited as language modeling evidence because it removes sequence mixing, but it directly tests whether the hourglass FFN parameterization remains stable in the absence of attention.
As shown in \autoref{tab:bigram}, the Hourglass FFN remains comparable to the conventional FFN under the same FFN parameter budget.

\begin{table*}[t!]
    \centering
    \caption{\textbf{Toy Bigram Language Model Results.} With self-attention removed, the Hourglass FFN remains comparable to the conventional FFN under a matched FFN parameter budget.}
    \label{tab:bigram}
    \resizebox{0.8\linewidth}{!}{

        \begin{tabular}{lc|cccc|cc}
            \toprule
            Model                 & FFN Size & $d_{\text{model}}$ & $d_{\text{h}}$ & $K$ & $L$ & Val Loss & Val PPL \\ % & Down Avg Acc & Down Avg PPL
            \midrule
            Conventional FFN      & 85M      & 768                & 3072           & 1   & 12  & 6.47     & 717.12  \\ % & 0.321 & 2.145
            Hourglass FFN ($K=4$) & 85M      & 854                & 690            & 4   & 12  & 6.467    & 715.09  \\ % & 0.325 & 2.144
            % \midrule
            % Hourglass FFN ($K=4$) & 85M      & 784                & 752            & 4   & 12  & 6.471    & 718.13  & 0.325        & 2.146        \\
            % Hourglass FFN ($K=5$) & 85M      & 768                & 614            & 5   & 12  & 6.472    & 719.00  & 0.323        & 2.145        \\
            % Hourglass FFN ($K=6$) & 85M      & 768                & 512            & 6   & 12  & 6.475    & 722.17  & 0.325        & 2.147        \\
            \bottomrule
        \end{tabular}
    }
\end{table*}

\subsection{FFN intermediate ratio at 403M.}
\label{app:fnn_intermediate_ratio_at_403M}
\begin{figure}[ht!]
    \centering
    \includegraphics[width=0.6\linewidth]{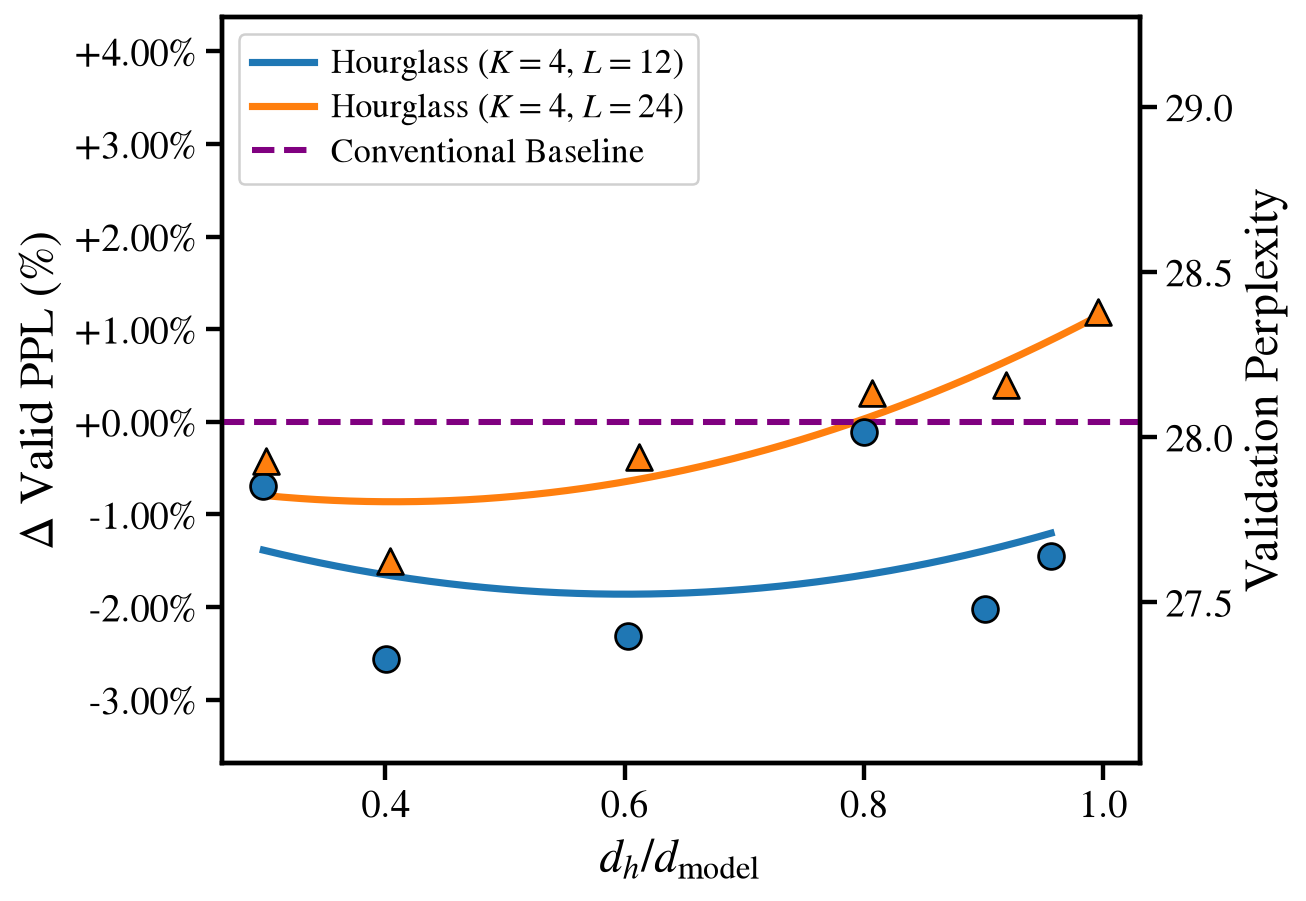}
    \caption{\textbf{$d_h/d_{\text{model}}$ search at 403M.} Validation perplexity across different $d_h/d_{\text{model}}$ ratios for Hourglass Transformers at 403M parameters. The $(K,L)=(4,12)$ curve (blue) remains below the conventional baseline across the sweep and is consistently lower than the $(K,L)=(4,24)$ curve (orange).}
    \label{fig:prop_ablation_curve}
\end{figure}

We fix the total model size to 403M parameters and sweep $d_h/d_{\text{model}}$ for Hourglass Transformers at two layer counts, as shown in \autoref{fig:prop_ablation_curve}.
Both $(K,L)=(4,12)$ and $(K,L)=(4,24)$ are competitive with or better than the conventional baseline ($L=24$, $d_h/d_{\text{model}}=4$) across most $d_h/d_{\text{model}}$ ratios.
Notably, the $(K,L)=(4,12)$ configuration---which uses only half the Transformer layers of the baseline---performs comparably or better than the $(K,L)=(4,24)$ variant across the sweep, suggesting that the freed parameters can be more effectively spent on wider hidden representations.

\subsection{Additional Hourglass FFN Ablations}

We conduct a series of ablation studies on the 113M hourglass FFN Transformer to investigate the impact of key architectural hyperparameters.
All experiments use the same dataset, tokenization, and training setup as described in
Section \ref{sec:experimental_setup}.

\begin{table}[ht!]
    \centering
    \caption{\textbf{Impact of increasing hourglass depth $K$.} Performance (validation loss and perplexity) improves as depth $K$ increases from 1 to 8. We fix $d_\text{model}=1032$, $d_h=418$, and $L=12$.}
    \label{tab:ablation_vary_k}
    \resizebox{0.75\linewidth}{!}{
        \begin{tabular}{l|ccc|cc}
            \toprule
            $K$ & Model Size & Attention Size & FFN Size & Val Loss & Val PPL \\
            \midrule
            1   & 67M        & 51M            & 16M      & 3.551    & 40.153  \\
            2   & 82M        & 51M            & 31M      & 3.489    & 37.632  \\
            4   & 113M       & 51M            & 62M      & 3.426    & 35.335  \\
            6   & 144M       & 51M            & 93M      & 3.391    & 34.051  \\
            8   & 175M       & 51M            & 124M     & 3.357    & 32.832  \\
            \bottomrule
        \end{tabular}
    }
    \vspace{-2mm}
\end{table}

\vspace{-1.5mm}
\subsubsection{Varying Hourglass FFN Depth \texorpdfstring{$K$}{K}}
\label{sec:ablation_vary_k}
\vspace{-2mm}
With $d_{\text{model}} = 1032$, $d_h = 418$, and $L = 12$ fixed, we vary $K \in \{1, 2, 4, 6, 8\}$ to study the effect of Hourglass FFN depth.
As shown in \autoref{tab:ablation_vary_k}, increasing $K$ consistently improves performance: validation perplexity drops from 40.153 at $K=1$ to 32.832 at $K=8$, with corresponding reductions in both training and validation loss.
The gains are most pronounced when moving from shallow configurations ($K=1$ or $K=2$) to moderate depth ($K=4$ or $K=6$), after which improvements become more incremental.

Notice that under this setting, larger $K$ also increases model size from 67M at $K=1$ to 175M at $K=8$, suggesting a trade-off between depth and efficiency.
This indicates that deeper Hourglass FFNs yield better performance in this sweep, while moderate $K$ values (e.g., $K=4$ or $K=6$) provide a strong accuracy--parameter trade-off for the main architecture design.

% \begin{table*}[ht!]
%     \centering
%     \caption{\textbf{Impact of varying $d_h$.} The validation perplexity increases gradually as the ratio decreases from 0.8 to 0.4. We fixed $d_\text{model}=1032$, $K=4$, and $L=12$.}
%     \label{tab:ablation_vary_dh}
%     \resizebox{\linewidth}{!}{
%     \begin{tabular}{lccccccccc}
%         \toprule
%         Model & Model Size & Attention Size & FFN Size & $d_{\text{model}}$ & $d_{\text{h}}$ & $K$ & $L$ & Val Loss & Val PPL \\
%         \midrule
%         Hourglass (ratio=0.8) & 175M & 51M & 124M & 1032 & 836 & 4 & 12 & 3.658 & 38.79 \\
%         Hourglass (ratio=0.6) & 144M & 51M & 93M  & 1032 & 627 & 4 & 12 & 3.686 & 39.89 \\
%         Hourglass (ratio=0.4) & 113M & 51M & 62M  & 1032 & 418 & 4 & 12 & 3.742 & 41.44 \\
%         Hourglass (ratio=0.2) & 82M  & 51M & 31M  & 1032 & 209 & 4 & 12 & 3.798 & 44.65 \\
%         Hourglass (ratio=0.1) & 66M  & 51M & 15M  & 1032 & 103 & 4 & 12 & 3.845 & 46.74 \\
%         \bottomrule
%     \end{tabular}
%     }
% \end{table*}

\begin{table}[ht!]
    %\vspace{-1mm}
    \centering

    \caption{\textbf{Impact of varying $d_h$.} The validation perplexity increases gradually as the ratio decreases from 0.8 to 0.1. We fixed $d_\text{model}=1032$, $K=4$, and $L=12$.}

    \label{tab:ablation_vary_dh}
    \resizebox{0.75\linewidth}{!}{
        \begin{tabular}{l|c|ccc|cc}
            \toprule
            $d_{\text{h}}$ & $d_h/d_{\text{model}}$ & Model Size & Attention Size & FFN Size & Val Loss & Val PPL \\
            \midrule
            836            & 0.8                    & 175M       & 51M            & 124M     & 3.355    & 32.788  \\
            627            & 0.6                    & 144M       & 51M            & 93M      & 3.384    & 33.747  \\
            418            & 0.4                    & 113M       & 51M            & 62M      & 3.426    & 35.335  \\
            209            & 0.2                    & 82M        & 51M            & 31M      & 3.500    & 38.075  \\
            103            & 0.1                    & 66M        & 51M            & 15M      & 3.541    & 39.741  \\
            \bottomrule
        \end{tabular}

    }
    \vspace{-1mm}
\end{table}
%\vspace{-1mm}
\subsubsection{Varying FFN Bottleneck Width \texorpdfstring{$d_h$}{d-h}}
\label{sec:ablation_vary_dh}

With $d_{\text{model}} = 1032$, $L = 12$, and $K = 4$ fixed, we vary $d_h/d_{\text{model}} \in \{0.8, 0.6, 0.4, 0.2, 0.1\}$ to study the effect of reducing the intermediate FFN dimension relative to the hidden dimension.
As shown in \autoref{tab:ablation_vary_dh}, decreasing the ratio leads to smaller model sizes from 175M at ratio $=0.8$ down to 66M at ratio $=0.1$ while performance degrades gradually.
Validation perplexity increases from 32.788 at ratio $=0.8$ to 39.741 at ratio $=0.1$, with the largest jumps occurring when the ratio drops below $0.4$.

These results characterize the trade-off between FFN width and model efficiency under a fixed architectural configuration. Reducing $d_h$ leads to substantial parameter savings in the FFN, while validation performance degrades gradually over a broad range of ratios. In particular, configurations with $d_h/d_{\text{model}}\ge 0.4$ maintain competitive performance despite significant reductions in model size. Below this range, performance degradation becomes more pronounced, indicating a transition where further FFN compression is less effective. Overall, the results identify a practical operating range for FFN width reduction in this small-scale hourglass setting.

% These results offer a measure of FFN redundancy. 
% The fact that we can reduce the width ratio to $\approx 0.4$ without significant performance loss (and often with gains when $K$ is increased) strongly suggests that standard FFNs are over-parameterized in the width dimension.
% The "wide" part of the narrow-wide-narrow convention serves as a safe but inefficient default; a narrower, deeper bottleneck is a computationally dense alternative.

\subsection{Training FLOP Accounting}
\label{app:train_flops}
We adapted the flop estimation reported by \citep{hoffmann2022trainingcomputeoptimallargelanguage,mcleish2025gemstones}. We report the train FLOPs for the forward and backward passes.

A python excerpt is as follows:
\begin{lstlisting}[language=Python]
def calculate_train_flops(
    d_model, d_model_attn, d_h, k_mlp_layers,
    n_layers, batch_size, seq_len, num_heads,
    use_qk_norm=True, vocab_size=100278,
):
    num_qheads = num_heads
    num_kvheads = 2 * num_qheads
    head_size = d_model_attn // num_heads
    d_h = d_h / 2  # SwiGLU splits d_h in half

    # === Forward Pass ===
    # Attention
    qkv_proj = batch_size * 3 * (2.0 * seq_len * d_model * d_model_attn)
    rope = batch_size * 3.5 * seq_len * (num_qheads + num_kvheads / 2) * head_size
    attention = qkv_proj + rope
    kq_logits = batch_size * seq_len * seq_len * d_model_attn
    softmax = batch_size * 3.0 * seq_len * seq_len * num_qheads
    attn_v = batch_size * 2.0 * seq_len * seq_len * d_model_attn
    attn_out = batch_size * 2.0 * seq_len * d_model * d_model_attn
    qk_norm = batch_size * 7.0 * (2 * d_model_attn) * seq_len if use_qk_norm else 0

    # MLP (K hourglass sub-layers)
    mlp_proj = k_mlp_layers * (batch_size * 6.0 * seq_len * d_model * d_h)
    mlp_act = k_mlp_layers * (batch_size * 10.0 * seq_len * d_h)

    # RMSNorm and residuals
    attn_norm = batch_size * 7.0 * d_model * seq_len
    attn_res = batch_size * d_model * seq_len
    mlp_norm = k_mlp_layers * (batch_size * 7.0 * d_model * seq_len)
    mlp_res = k_mlp_layers * (batch_size * d_model * seq_len)
    total_norm = attn_norm + mlp_norm
    total_res = attn_res + mlp_res

    # LM head
    lm_head = batch_size * (2 * seq_len * d_model * vocab_size)

    layer_fwd = (attention + kq_logits + softmax + attn_v + attn_out
                 + qk_norm + mlp_proj + mlp_act + total_norm + total_res)
    final_norm = batch_size * 7.0 * d_model * seq_len
    total_fwd = n_layers * layer_fwd + final_norm

    # === Backward Pass ===
    layer_attn_bwd = (2.0 * attention
        + 2.5 * (kq_logits + softmax + attn_v)
        + 2.0 * attn_out + 2.0 * qk_norm)
    layer_nonattn_bwd = 2.0 * (mlp_proj + mlp_act + total_res + total_norm)
    total_bwd = (n_layers * (layer_attn_bwd + layer_nonattn_bwd)
                 + 2.0 * final_norm)

    # === Total training FLOPs ===
    total = total_fwd + total_bwd + lm_head + 2.0 * lm_head
    return total
\end{lstlisting}

\section{Inference Cost Analysis}
\label{app:inference_cost_analysis}
\vspace{-2mm}

% The paper studies how to allocate representation width and FFN capacity while keeping the feature dimension of the sequence-mixing operation, $d_{\text{attn}}$, matched to the conventional baseline for fair comparison. It does not propose a replacement for MHA, GQA, or MLA. These mixers change the KV width within each attention layer, whereas Hourglass changes the number of attention layers $L$. Therefore, the two mechanisms act on different factors of the inference cost.

Formally speaking, let $S$ be the cached context length, $\beta$ the bytes per cached element ($\beta=2$ for BF16), and $g$ the per-layer KV-compression factor for different sequence mixers ($g=1$ for MHA, $g=h/n_{\text{kv}}$ for GQA, and $g=2d_{\text{attn}}/(r_{\text{KV}}+d_{\text{rope}})$ for the MLA cache layout considered here). The total KV-cache memory and the KV bytes read for one decoded token are
$$
M_{\text{KV}}(S;g)=\frac{2\beta L S d_{\text{attn}}}{g}
=\frac{4L S d_{\text{attn}}}{g}\ \text{bytes in BF16}.
$$

For MHA and GQA, the sequence-length-dependent QK and AV operations require approximately
$$
F_{\text{attn}}(S)\simeq 4L S d_{\text{attn}}\ \text{FLOPs}
$$
per decoded token. Equivalently, the leading total decode costs in terms of compute and memeory can be written as
$$
F_{\text{decode}}(S)\simeq F_{\text{fixed}}+4L S d_{\text{attn}},
\qquad
M_{\text{decode}}(S;g)\simeq M_{\text{weights}}+
\frac{4L S d_{\text{attn}}}{g},
$$
where the fixed terms include the FFN and projection operations and the model-weight traffic and $BW_{\text{eff}}$ is the effective GPU memory bandwidth. MLA has additional implementation-dependent projection terms, but its context-dependent cache traffic remains linear in $L$, $S$, and its compressed KV width. 

In the memory-bound regime, the corresponding first-order latency model is
$$
t_{\text{decode}}(S;g)
=t_{\text{fixed}}+\frac{4L S d_{\text{attn}}}{g\,\mathrm{BW}_{\text{eff}}},
$$
where $t_{\text{fixed}}$ collects context-independent work such as weight access and kernel launches. Because the compared models match $d_{\text{attn}}$ and use the same mixer, their relative KV-cache size is
\begin{equation}
    \label{eq:kv_save}
    \frac{M_{\text{KV,hg}}}{M_{\text{KV,conv}}}
    =\frac{L_{\text{hg}}}{L_{\text{conv}}},
\end{equation}
independent of $g$. Thus Hourglass reduces both the attention FLOPs and KV traffic associated with the removed layers: by $50.0\%$ at 906M, $35.7\%$ at 3B, and $31.3\%$ at 8B. 

As to the latency speed up, it can be written as
\begin{equation}
    \label{eq:speedup}
    \operatorname{speedup}(S;g)=\frac{c_{\text{fixed,conv}}+L_{\text{conv}}S }{c_{\text{fixed,hg}}+L_{\text{hg}}S},
\end{equation}
where we simplify the expression with $c_\text{fixed}=t_{\text{fixed}}(g\mathrm{BW}_{\text{eff}})/(4d_{\text{attn}})$. The $t_{\text{fixed}}$ is hard to quantify analytically due to consideration of the kernel operations and the sequential operations across layers. Nevertheless, as the sequence length increases, $LS$ term should dominate, therefore approximating $\text{speedup}(S;g)\approx\frac{L_{\text{conv}}}{L_{\text{hg}}}$. This matches what is observed empirically for MHA at 64k context length, where speedups are achieved by $1.93\times$ at 906M, $1.52\times$ at 3B, and $1.43\times$ at 8B, respectively.

\subsection{Empirical Comparison Across Sequence Mixers}
\label{app:mixer_decode_latency}

The model above predicts that changing the sequence mixer moves $g$ but leaves both the KV-cache ratio and the asymptotic speedup at $L_{\text{conv}}/L_{\text{hg}}$, shifting only the context length at which the asymptote is approached.
We test this at 906M by swapping the mixer in both arms of the matched pair while holding $d_{\text{attn}}=1536$, $L_{\text{conv}}=24$, and $L_{\text{hg}}=12$ fixed, and measuring single-token decode latency at 32k, 64k, and 128k.
The GQA configuration uses $h=16$ query heads over $n_{\text{kv}}=4$ key--value heads ($g=4$), and the MLA configuration uses $r_{\text{KV}}=256$ and $d_{\text{rope}}=32$ ($g\approx10.7$).

\paragraph{Inference setting.}
Measurements use an NVIDIA RTX A6000 with batch size 1 and BF16, timing a single decoded token against a fixed cache of the stated length, with 5 warm-up and 20 timed iterations pinned to an otherwise idle device.
Both architectures run through one identical code path per mixer, with the same attention backend selection and causal-mask handling and no architecture-specific kernel tuning.
To avoid an $O(S^2)$ prefill at every length, we build the cache directly: a short prefill records each layer's exact key/value structure, which is then rebuilt at the target sequence length and decoded against once.
The setup has two limitations: all measurements use batch size 1, and the decode kernel is not optimized, both of which reduce the observed speedup at a fixed context relative to the asymptote.

% This was validated to within $2\%$ of a full real-prefill measurement at 32k for every mixer.
% These measurements use different hardware and a different inference codebase from Figure~\ref{fig:latency_comparison}, so absolute latencies are not comparable across the two; only the trend within this table is meaningful.

\begin{table}[htb]
    \centering
    \caption{\textbf{Decode speedup and KV-cache reduction across sequence mixers at 906M.} Speedup is conventional latency divided by Hourglass latency for a single decoded token, so higher is faster; the two arms differ only in layer count ($L=24$ vs.\ $12$). For every mixer the speedup rises toward the layer-count ratio $L_{\text{conv}}/L_{\text{hg}}=2$ as context grows, and does so later for the mixers with larger $g$. The KV-cache reduction is exactly $1-L_{\text{hg}}/L_{\text{conv}}$ for every mixer and every context length, as the memory model predicts.}
    \label{tab:mixer_decode_latency}
    \setlength{\tabcolsep}{6pt}
    \renewcommand{\arraystretch}{1.05}
    \footnotesize
    \resizebox{0.5\linewidth}{!}{
        \begin{tabular}{l|ccc|c}
            \toprule
                                  & \multicolumn{3}{c|}{Decode speedup $\uparrow$} & KV-cache                                            \\
            \cmidrule(lr){2-4}
            Sequence mixer        & 32k                                            & 64k          & 128k         & reduction $\uparrow$ \\
            \midrule
            MHA ($g=1$)           & $1.93\times$                                   & $1.96\times$ & $1.98\times$ & $50.0\%$             \\
            GQA ($g=4$)           & $1.87\times$                                   & $1.93\times$ & $1.97\times$ & $50.0\%$             \\
            MLA ($g\approx10.7$)  & $1.87\times$                                   & $1.94\times$ & $1.97\times$ & $50.0\%$             \\
            \bottomrule
        \end{tabular}
    }
\end{table}

\autoref{tab:mixer_decode_latency} matches the inference memory model at \autoref{eq:kv_save}.
The KV-cache reduction is $50.0\%$ for every mixer at every length, confirming that the relative memory saving is independent of $g$.
Latency behaves as predicted as well.
At a fixed context, mixers with larger $g$ sit further from the asymptote, because a smaller cache lets the context-independent term $t_{\text{fixed}}$ occupy a larger share of the step; by 128k every mixer is within $0.03\times$ of the layer-count ratio and the layer ratio in \autoref{eq:speedup} dominates.

% \input{src/figure_layout_proposal}
% \input{src/analysis_cod}

% \newpage
% \input{checklist.tex}

\end{document}